\pdfoutput=1
\RequirePackage{fix-cm}
\documentclass[twocolumn]{svjour3}
\smartqed

\usepackage[T1]{fontenc}
\usepackage{times}
\usepackage{amsmath}
\usepackage{amssymb}
\usepackage{amsfonts}
\usepackage{bm}
\usepackage{siunitx}
\usepackage{graphicx}
\usepackage[table]{xcolor}
\usepackage{multirow}
\usepackage{diagbox}
\usepackage{array}
\usepackage{booktabs}
\usepackage{tabularray}
\usepackage{subcaption}
\usepackage{float}
\usepackage{natbib}
\usepackage[colorlinks,
            linkcolor=green,
            anchorcolor=blue,
            citecolor=blue]{hyperref}

\makeatletter
\def\makeheadbox{}
\makeatother
\begin{document}
\begin{sloppypar}

\title{Hallo4D: Multi-Modal Hallucination Mitigation for Consistent Spatio-Temporal Generation}

\titlerunning{Hallo4D: Multi-Modal Hallucination Mitigation for Consistent Spatio-Temporal Generation}
\authorrunning{H. Wang et al.}

\author{Hongbo~Wang$^{1,2}$ \and Huaibo~Huang$^{1,2}$ \and Jie~Cao$^{1,2}$ \and Jin~Liu$^{1,3}$ \and Haoyang~Tong$^{1,2}$ \and Ran~He$^{1,2}$}

\institute{
           {$^1$} MAIS\&NLPR, Institute of Automation, Chinese Academy of Science, Beijing, 100190, China \\
           {$^2$} University of Chinese Academy of Sciences, Beijing, 101408, China \\
           {$^3$} Shanghaitech University, Shanghai, 201210, China \\
           \email {wanghongbo2024@ia.ac.cn, \{huaibo.huang, jie.cao\}@cripac.ia.ac.cn, liujin2@shanghaitech.edu.cn, tonghaoyang22@mails.ucas.ac.cn, rhe@nlpr.ia.ac.cn} \\
}

\date{}
\maketitle

\begin{abstract}
While recent progress in 3D generation has enabled impressive visual synthesis, most existing methods still primarily rely on 2D diffusion-based supervision without mechanisms for enforcing geometric consistency, often resulting in spatial hallucinations such as duplicated structures or misaligned geometry. 
These challenges intensify in 4D generation, where maintaining consistency across viewpoints and temporal progression is substantially more difficult and often leads to temporal artifacts such as jitter, identity flicker, and structural drift.
To address these limitations, we present \textbf{Hallo4D}, a unified and model-agnostic framework for mitigating spatiotemporal hallucinations in both 3D and 4D content generation. Hallo4D introduces a generation-detection-correction paradigm that leverages the reasoning capabilities of large multimodal language models (LMMs) to locate and summarize spatial and temporal inconsistencies from multi-view, multi-frame renderings. To prevent compounding errors from single-pass edits and ensure robust geometric fidelity, these insights guide a consensus-driven image-space consistency optimization, where an LMM selector evaluates multiple candidate corrections via multi-model voting. This process is achieved without requiring retraining or architectural modifications.
To further enhance temporal consistency, Hallo4D incorporates a motion-saliency-driven keyframe sampling strategy based on optical flow, enabling more targeted and efficient refinement. The framework also includes an LMM-guided initialization scheme and an attention-based appearance alignment module to improve early optimization and cross-view fidelity.  Additionally, we address exposure instability with two losses: Contrastive Semantic Exposure Alignment (CSEA), a foreground-masked contrastive objective that favors well-exposed semantics while penalizing over- and under-exposure, and a log-dynamic-range (LDR) loss that regularizes luminance contrast. Together with union-of-frusta visibility pruning to remove out-of-view clutter and reduce pseudo under-exposure, these additions mitigate exposure-driven collapse under non-frontal views.
Extensive experiments demonstrate that Hallo4D consistently outperforms strong baselines across diverse generation settings, offering a scalable and generalizable solution for consistency-aware 3D and 4D content generation. Additional visualizations are available on our project page: \url{https://wafer-bob.github.io/Hallo3D-4D/}.

\end{abstract}

\section{Introduction}
\label{sec:introduction}

\begin{figure*}[t]
    \centering
    \includegraphics[width=0.95\linewidth]{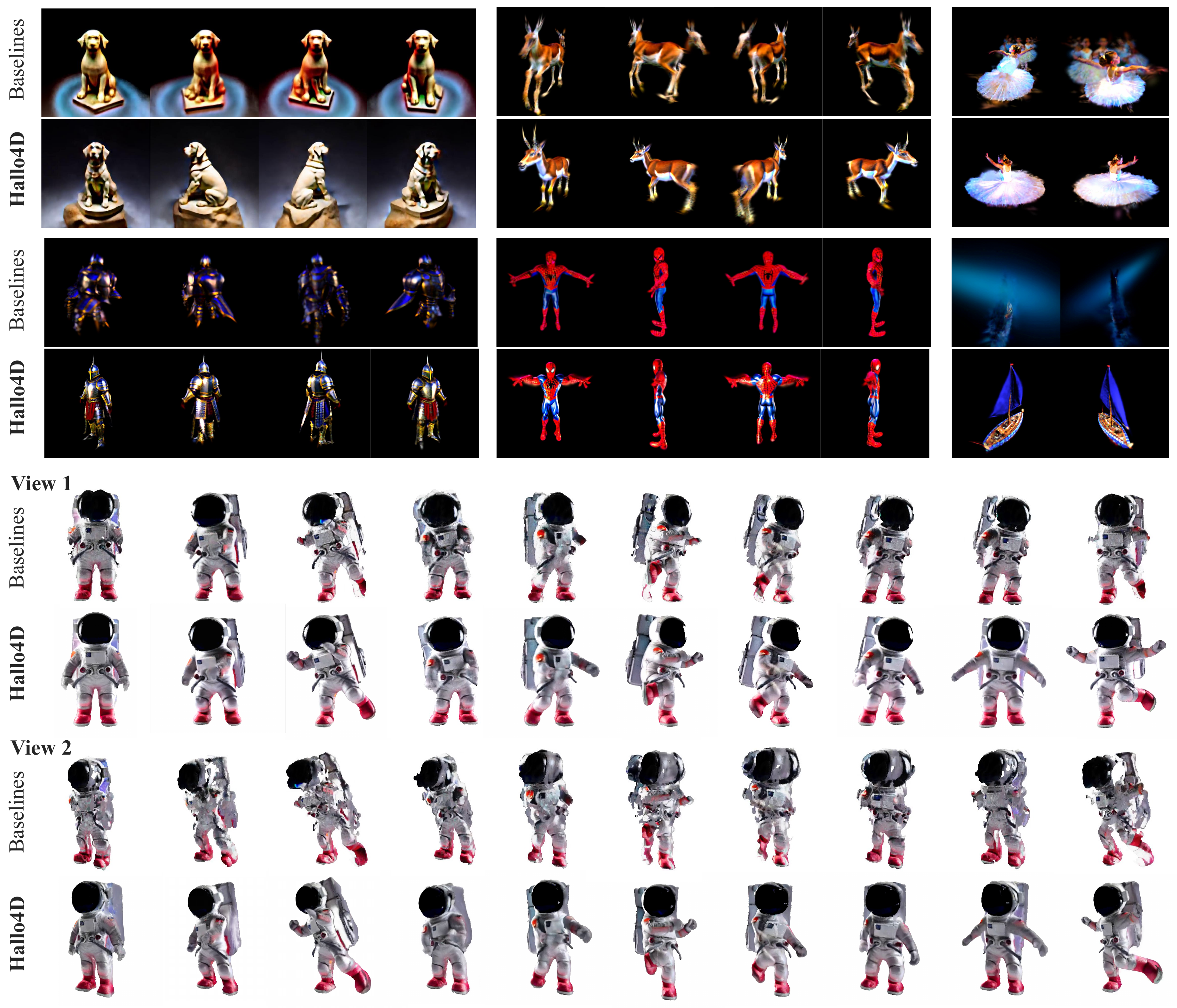}
    \caption{\textbf{Comparison of 3D and 4D generation results between Hallo4D and the baseline.} Our method achieves significant improvements in spatiotemporal consistency.}
    \label{fig:teaser}
\end{figure*}

Three-dimensional content generation has become a foundational capability for AI-driven visual understanding and simulation. With the advent of diffusion-based supervision, recent methods can synthesize high-fidelity 3D assets from minimal input, using text prompts~\citep{pooleDreamFusionTextto3DUsing2022, junShapEGeneratingConditional2023}, single-view images~\citep{liuZero1to3ZeroshotOne2023}, or sparse multiview data~\citep{jiangEfficient3DiMLearningGeneralizable2023, shiMVDreamMultiviewDiffusion2024}. A particularly successful direction leverages 2D diffusion models as supervisory priors by training 3D representations to match the distribution of rendered views with that of a pre-trained image diffusion model. This is operationalized through Score Distillation Sampling (SDS)~\citep{pooleDreamFusionTextto3DUsing2022}, which iteratively updates 3D parameters to minimize view-level discrepancies against the learned 2D prior.
While remarkably effective, these approaches rely exclusively on view-dependent 2D supervision, which often leads to hallucinated or duplicated structures in unobserved views—a failure mode commonly referred to as the Janus problem\citep{armandpourReimagineNegativePrompt2023}. This issue reflects a core weakness of current pipelines: the absence of mechanisms to explicitly model and enforce geometric consistency across viewpoints. Several follow-up works attempt to address this by incorporating 3D-aware priors\citep{zhaoEfficientDreamerHighFidelityRobust2023, liuSyncDreamerGeneratingMultiviewconsistent2024} or explicitly enforcing multiview coherence~\citep{yangConsistNetEnforcing3D2023}, but such solutions remain constrained by limited supervision and still face challenges in achieving robust generation.

Importantly, while existing efforts primarily focus on improving spatial consistency within static 3D content, real-world scenarios demand coherence that extends across both space and time. Applications such as animation~\citep{xuMagicAnimateTemporallyConsistent2024a} and tracking~\citep{wuSingleModelAnyModalityVideo2024} require modeling dynamic geometry that evolves continuously over time, naturally calling for 4D content generation to capture the spatiotemporal continuity of the physical world.
Transitioning from static to dynamic geometry, however, introduces a new set of challenges that current 4D generation approaches have yet to resolve in a unified way. Some methods build upon dynamic NeRF frameworks~\citep{pumarolaDNeRFNeuralRadiance2020, liNeural3DVideo2022} by decoupling the canonical representation from the deformation field to model motion over time, achieving high-quality 4D content. Yet, their reliance on latent-space encoding often compromises geometric consistency across frames, leading to temporal artifacts such as jittering and ghosting~\citep{zhengUnifiedApproachTextand2024, jiangConsistent4DConsistent360deg2023}.
Meanwhile, other approaches adopt explicit neural structures~\citep{caoHexPlaneFastRepresentation2023} to model deformation, improving inference efficiency and robustness to local non-rigid motion. Nevertheless, maintaining high-fidelity textures and ensuring stable spatiotemporal consistency across complex dynamic scenes remains highly challenging~\citep{renDreamGaussian4DGenerative4D2023, bahmani4DfyTextto4DGeneration2024, singerTextto4DDynamicScene2023}.

To alleviate the above limitations, we propose \textbf{Hallo4D}, a spatiotemporal consistency optimization framework for detecting and mitigating hallucinations in both 3D and 4D content generation. Central to our approach is the observation that LMMs exhibit strong spatial reasoning capabilities, allowing them to identify and summarize inconsistencies that emerge across both views and timesteps in dynamic scenes. Leveraging this capability, Hallo4D implements a novel generation-detection-correction paradigm that operates directly on multi-view, multi-frame renderings—without requiring any additional training data or task-specific supervision. 
By combining this paradigm with a consensus-driven diffusion-based 2D editing technique, which evaluates multiple candidate corrections to prevent error accumulation, our framework can be seamlessly applied to a wide range of existing 3D and 4D generation models without modifying their architectures or retraining procedures.
Beyond consistency correction, Hallo4D further improves appearance quality in 3D and temporal coherence in 4D. We introduce an attention-based alignment module to enhance cross-view texture consistency, and a motion-saliency-driven keyframe sampling strategy to improve temporal stability and efficiency. 
In addition, we address exposure instability with two exposure-aware losses, Contrastive Semantic Exposure Alignment (CSEA) and a log-dynamic-range (LDR) loss, and apply union-of-frusta visibility pruning to suppress out-of-view clutter and reduce pseudo under-exposure; these additions mitigate exposure-driven collapse under non-frontal views.
These components together support reliable and high-quality generation across diverse scenarios.

Specifically, for 3D generation, we first render images from multiple viewpoints and construct a 3D-inconsistency-aware inquiry to query the LMM. Through standardized responses, the LMM identifies cross-view inconsistencies and generates enhanced negative prompts, which are then used to guide targeted image-space optimization. This process leverages a Multi-view Appearance Alignment strategy, where a focal view supplies key and value features to a cross-attention mechanism, ensuring texture consistency across different viewpoints.
For 4D generation, we introduce three additional mechanisms to address the more severe spatiotemporal challenges. 
\textit{First, an LMM-guided initialization stage.} This stage jointly analyzes static 3D assets and animated sequences to detect cross-view and cross-timestep inconsistencies. The identified hallucinations are transformed into targeted initialization losses, ensuring that early optimization is guided by stable supervision and reducing the risk of compounding errors in subsequent stages.
\textit{Second, OF-Range for keyframe sampling.} This optical-flow-based strategy prioritizes frames with salient motion while maintaining temporal dispersion. By focusing optimization on dynamically informative frames and avoiding redundant updates on static regions, OF-Range improves temporal coherence and enhances computational efficiency during 4D training.
\textit{Third, an exposure-aware semantic alignment module.} This module prevents view-dependent collapse by introducing two complementary losses: Contrastive Semantic Exposure Alignment (CSEA), a foreground-masked contrastive objective that aligns with well-exposed semantics while penalizing over- and under-exposure, and a log-dynamic-range (LDR) term that regularizes contrast in the log-luminance domain. Together with union-of-frusta visibility pruning to remove out-of-view clutter and alleviate pseudo under-exposure, this module stabilizes optimization under non-frontal views and improves overall robustness.

Across both 3D and 4D stages, identified inconsistencies are corrected through a Prompt-Enhanced Re-consistency module, which uses diffusion-based image-space editing conditioned on enhanced negative prompts. 
Across both 3D and 4D stages, identified inconsistencies are corrected through a \textit{Consensus-Driven Re-consistency via Prompt Enhancement} module. Instead of relying on a single deterministic edit that may introduce compounding errors, this module generates multiple candidate corrections guided by enhanced negative prompts, and employs an LMM Consensus Selector to determine the optimal geometrically faithful result via multi-model voting.
To ensure theoretical validity, we establish an image-space reformulation of SDS that supports these corrections without disrupting the underlying training dynamics.
Extensive experiments show that Hallo4D consistently outperforms strong baselines in spatial, temporal, and perceptual quality metrics, demonstrating its robustness and broad applicability across diverse 3D and 4D generative tasks. 

\textbf{Our Contributions.} We summarize our contributions as follows:
\begin{itemize}
\item We propose Hallo4D, a unified spatiotemporal consistency framework for detecting and mitigating hallucinations in 3D and 4D generation. Hallo4D operates without additional data or retraining, ensuring broad applicability across diverse models.
\item We develop a novel generation-detection-correction paradigm that leverages LMMs for inconsistency detection and a consensus-driven image-space optimization for correction. We demonstrate that LMMs can not only infer spatial structures and diagnose multi-view and temporal inconsistencies but also act as an effective ensemble selector to ensure robust geometric fidelity via multi-model voting.
\item We design a 4D initialization scheme and an optical-flow-based motion saliency sampling strategy to improve early-stage optimization and enhance temporal stability in 4D generation. Additionally, we introduce two exposure-aware losses, together with union-of-frusta visibility pruning to reduce exposure instability under non-frontal views
\item Extensive experiments show that Hallo4D consistently improves visual quality and spatiotemporal consistency over strong baselines across various 3D and 4D tasks.
\end{itemize}

\textbf{Differences from Our Prior Work \textit{Hallo3D}~\citep{wangHallo3DMultimodalHallucination2024}} published in NeurIPS 2024.
This paper presents Hallo4D, a substantial extension of~\citep{wangHallo3DMultimodalHallucination2024}, with major advances in scope, modeling capabilities, and theoretical grounding:

\begin{itemize}
    \item \textbf{Methodological Enhancement:} Hallo4D extends spatial-only correction to unified spatiotemporal optimization. Crucially, we upgrade Hallo3D's deterministic single-pass editing to a Consensus-Driven Prompt-Enhanced Re-consistency mechanism, mitigating compounding errors via multi-model voting. Coupled with LMM-guided initialization and motion-saliency-based keyframe sampling, this ensures high-quality, temporally coherent 4D generation. Furthermore, two novel exposure-aware losses (CSEA and LDR) are introduced to stabilize illumination and reduce pseudo under-exposure in non-frontal views. To our knowledge, this is the first plug-and-play, model-agnostic consistency optimization strategy for 4D generation.

    \item \textbf{Functional Generalization:} The role of LMMs is further generalized in both scope and depth of application. Originally used to detect spatial inconsistencies in 3D outputs, LMMs are now shown to effectively identify temporal inconsistencies in 4D sequences, demonstrating their capacity to reason over both multi-view geometry and cross-frame dynamics. In addition, we extend their usage beyond post-hoc analysis by incorporating LMM-derived signals into the early optimization phase, enabling broader participation in consistency-aware generation.
    
    \item \textbf{Theoretical Grounding and Expanded Evaluation:} We provide new theoretical analysis that explains why image-space consistency methods can effectively transfer to 3D and 4D geometry refinement, offering a stronger conceptual basis for our correction strategy. In addition, we broaden the empirical evaluation to include high-quality baselines not previously covered~\citep{wangProlificDreamerHighFidelityDiverse2023,longWonder3DSingleImage2023}, demonstrating the generality and robustness of the proposed framework across diverse generative settings.
\end{itemize}

\section{Related Work}
\subsection{3D Content Generation}
The advent of diffusion models has revolutionized text-to-3D generation, enabling the synthesis of high-fidelity 3D assets from textual descriptions. DreamFusion \citep{pooleDreamFusionTextto3DUsing2022} pioneered this field by leveraging Score Distillation Sampling (SDS) to optimize 3D structures, integrating MipNeRF 360 \citep{barronMipNeRF360Unbounded2022} for neural rendering and Imagen \citep{sahariaPhotorealisticTexttoImageDiffusion2022} for high-quality text-to-image generation.
While NeRF-based approaches \citep{mildenhallNeRFRepresentingScenes2022, lorraineATT3DAmortizedTextto3D2023, zhouDreamPropellerSuperchargeTextto3D2023, wangProlificDreamerHighFidelityDiverse2023, liInstant3DInstantTextto3D2023,gaoGraphDreamerCompositional3D2023, linMagic3DHighResolutionTextto3D2023}  improve photorealism and lighting effects, their high computational cost limits scalability. To address this, 3D Gaussian Splatting (3DGS) \citep{kerbl3DGaussianSplatting2023} proposed an efficient alternative by representing 3D scenes with optimizable Gaussians, enabling real-time rendering. Subsequent refinements \citep{chenTextto3DUsingGaussian2023, yiGaussianDreamerFastGeneration2023, diHyper3DGTextto3D2025} have improved compositionality and efficiency. On the optimization side, CoGrad3D~\citep{tongCoGrad3DSpatiallyCoupledTimestep2026} couples spatially-aware timestep scheduling with orthogonal gradient fusion across viewpoints, further improving view consistency and texture fidelity.
Beyond conventional methods, GAN-based models \citep{attaikiGANFusionFeedForwardTextto3D2024, li3DGOI3DGAN2024} integrate GANs with diffusion models for fast, eliminating iterative optimization. Meanwhile, scene-level methods \citep{zhouGALA3DTextto3DComplex2024, liDreamScene3DGaussianbased2024} improve structural and semantic coherence, expanding text-to-3D synthesis capabilities.

Images from specific viewpoints provide stronger visual consistency for 3D generation. Image-based methods \citep{tangDreamGaussianGenerativeGaussian2023, fuCOLMAPFree3DGaussian2023, liuGhostShellExpressive2023, tangMakeIt3DHighFidelity3D2023a, albaharSingleImage3DHuman2023, qianMagic123OneImage2023, longWonder3DSingleImage2023} often outperform text-based approaches by leveraging accurate multi-view supervision. 3D-aware image generation techniques \citep{xiang3DawareImageGeneration2023, dengGRAMGenerativeRadiance2022} enhance rendering consistency across viewpoints but still face challenges due to training data scarcity. Recent diffusion-based models \citep{junShapEGeneratingConditional2023, nicholPointESystemGenerating2022} incorporating 3D priors have improved geometric realism, reducing perceptual errors and advancing the fidelity of image-to-3D synthesis. More recently, feed-forward generative models built on structured 3D latents~\citep{xiangStructured3DLatents2025} have substantially improved the scalability and fidelity of direct 3D asset generation.

\subsection{4D Content Generation}
Generating 4D content~\citep{miaoAdvances4DGeneration2025}, which extends 3D synthesis by introducing temporal dynamics, requires balancing motion realism, efficiency, and control to create coherent, high-fidelity dynamic scenes. Early efforts built on NeRF-based models \citep{duNeuralRadianceFlow2021, pumarolaDNeRFNeuralRadiance2020} or theoretical methods based on geometry \citep{laga4DAtlasStatistical2021}, but these struggled with high computational costs and temporal inconsistencies. More recent diffusion-based techniques \citep{bahmani4DfyTextto4DGeneration2024, bahmaniTC4DTrajectoryConditionedTextto4D2024, lingAlignYourGaussians2024} introduced score distillation and trajectory conditioning, significantly improving motion coherence and structure. Parallel to this, video-driven methods \citep{chuDreamScene4DDynamicMultiObject2024, jiangConsistent4DConsistent360deg2023, yangNotAllFrame2025, panEfficient4DFastDynamic2025} refined multi-object interactions, viewpoint consistency, and frame-wise feature disentanglement, addressing challenges in scene complexity. Meanwhile, physics-aware approaches \citep{zhangPhysDreamerPhysicsBasedInteraction2024, huangDreamPhysicsLearningPhysicsBased} incorporated real-world constraints, enhancing realism in object dynamics.

As generative models advanced, the demand for efficient scene representations grew. Traditional volumetric techniques were computationally expensive, leading to the rise of Gaussian splatting \citep{wu4DGaussianSplatting2024, linGaussianFlow4DReconstruction2024, renDreamGaussian4DGenerative4D2023, zengSTAG4DSpatialTemporalAnchored2024, lingAlignYourGaussians2024}, enabling real-time rendering while preserving fidelity. Diffusion-driven motion synthesis \citep{blattmannStableVideoDiffusion2023, sunDimensionXCreateAny2024} further improved scalability, while LLM-guided compositional approaches \citep{xuComp4DLLMGuidedCompositional2024} added structural priors for greater scene control. Beyond efficiency, structured motion control remains crucial. Hybrid representations \citep{liDreamMesh4DVideoto4DGeneration2024, xieSV4DDynamic3D2024} enhanced multi-view consistency and sparse motion control, while high-resolution 4D synthesis \citep{li4K4DGenPanoramic4D2024} improved detail retention in large-scale environments, enhancing adaptability and realism in complex dynamic settings. More recently, multi-view video diffusion models such as CAT4D~\citep{wuCAT4DCreateAnything2025} and feed-forward 4D generative frameworks~\citep{chen4DNeXFeedForward4D2025} enable direct 4D creation from monocular inputs, further broadening the applicability of dynamic content generation.

\begin{figure*}[!t]
  \centering
  \includegraphics[width=.95\textwidth]{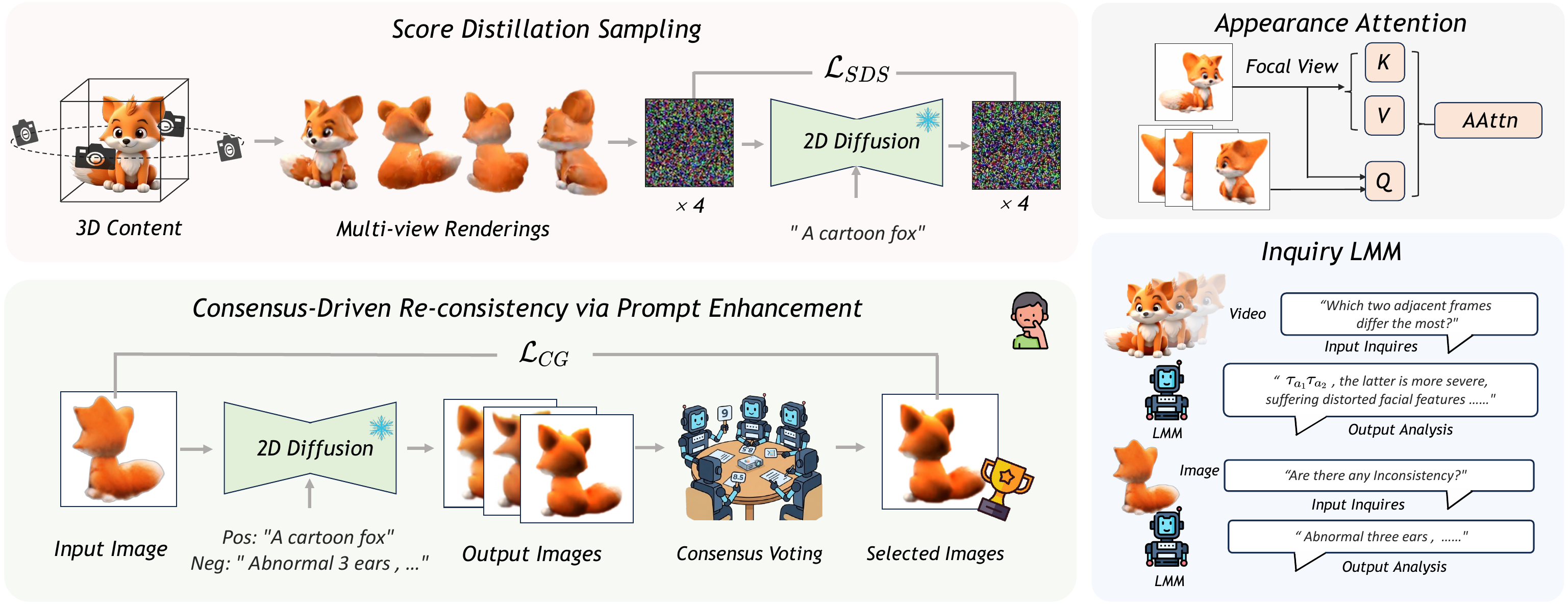}
  \caption{\textbf{Pipeline overview of the 3D consistency optimization stage.} We jointly optimize our model using $\mathcal{L}_{\rm{SDS}}$ and $\mathcal{L}_{\rm{CG}}$. For $\mathcal{L}_{\rm{SDS}}$, a focal view is selected based on camera pose and used as keys and values to align all views via attention. For $\mathcal{L}_{\rm{CG}}$, hallucinations are identified using LMMs and translated into enhanced negative prompts. These prompts guide the generation of multiple candidate corrections, from which an MLLM consensus selector determines the optimal result via multi-model voting to enforce re-consistency.}
  \label{fig:pipeline_3d}
\end{figure*}

\subsection{Spatio-temporal Consistency Enhancement}
Ensuring consistency is fundamental for generating coherent 3D and 4D content. In 3D generation, the primary focus lies in maintaining multi-view consistency, however, a key challenge arises from the reliance on 2D diffusion models, which often introduces viewpoint-specific hallucinations.
To mitigate this, fine-tuning approaches \citep{shiMVDreamMultiviewDiffusion2024, zhaoEfficientDreamerHighFidelityRobust2023, yangConsistNetEnforcing3D2023, liuSyncDreamerGeneratingMultiviewconsistent2024, jiangEfficient3DiMLearningGeneralizable2023} incorporate 3D constraints into diffusion models, improving consistency for 3D subjects \citep{ruizDreamBoothFineTuning2023, rajDreamBooth3DSubjectDrivenTextto3D2023}, transparent objects \citep{zhangTransparentImageLayer2024}, and multiple viewpoints \citep{seoLet2DDiffusion2023, jiDiffPanoScalableConsistent2026}. Beyond fine-tuning, prompt optimization \citep{hongDebiasingScoresPrompts2023, armandpourReimagineNegativePrompt2023}, object-level self-correction \citep{sunMarmotObjectLevelSelfCorrection2026}, and instance-aware preference optimization \citep{sunFineGrainedAttributionInstanceAware2026} refine alignment, while geometric corrections \citep{liuSherpa3DBoostingHighFidelity2023} and video-inspired frameworks \citep{voletiSV3DNovelMultiview2024} offer alternative solutions. Additionally, methods like \citet{liu3DGSenhancerEnhancingUnbounded2024, liangLucidDreamerHighFidelityTextto3D2023} enhance consistency and fidelity across dynamic and complex 3D content, and recent think-driven approaches \citep{jiaoThinkGenGeneralizedThinking2026, guoThinkingWhileGeneratingInterleaving2026, liuThinkThenGenerateStructural2026, fanMOC3DManifoldOrderConsistency2026} leverage the reasoning capabilities of large multimodal models to enforce structural consistency in visual and 3D generation. However, these approaches still face challenges with non-orthogonal viewpoints and adaptability across different architectures.

Building on multi-view alignment in 3D, recent efforts in 4D generation increasingly focus on ensuring temporal consistency across frames.
Techniques such as spatiotemporal anchoring and feature synchronization~\citep{zengSTAG4DSpatialTemporalAnchored2024, grammatikopoulouSpatiotemporalNetworkVideo2023} improve frame coherence and enable more consistent motion generation.
Meanwhile, temporal-consistent diffusion models and fast spatiotemporal optimization strategies~\citep{zhouUpscaleAVideoTemporalConsistentDiffusion2024, liangDiffusion4DFastSpatialtemporal2024} reduce flickering artifacts.
Further advances target long-range temporal correspondence and deformation minimization~\citep{yangFrescoSpatialTemporalCorrespondence2024, ouyangCoDeFContentDeformation2024}, while others~\citep{xuMagicAnimateTemporallyConsistent2024, wangCOVEUnleashingDiffusion2024} focus on producing smoother and more stable motion trajectories.
Progress in motion tracking and flow refinement~\citep{chenMapTrackerTrackingStrided2024, liangFlowVidTamingImperfect2024} has also contributed to enhancing the robustness of dynamic content generation.
While most existing methods focus on addressing consistency issues within specific models, efforts toward improving generality~\citep{qiuPnP3DPlugandPlay3D2023, wangVideo4DGenEnhancingVideo2025, liuFree4DTuningfree4D2025} remain confined to either 3D or 4D generation within limited domains.
In contrast, our proposed Hallo4D leverages LMMs and image-space optimization to achieve scalable and efficient consistency enhancement across both 3D and 4D generation tasks.

\section{Preliminaries} \label{sec:3.1_pre}
\subsection{Diffusion Models}
Diffusion models~\citep{hoDenoisingDiffusionProbabilistic2020} has a forward diffusion process with diffusion steps from $0$ to $T$, which degrades the original sample $\mathbf{x}_0$ into pure noise $\mathbf{x}_T$ over $T$ steps via:
\begin{gather}
    \label{eq:forward_ddpm}
    \mathbf{x}_t = \sqrt{\alpha_{{t}}}\mathbf{x}_0 + \sqrt{1-\alpha_{{t}}}\boldsymbol{\epsilon}, \quad \boldsymbol{\epsilon
    } \sim \mathcal{N}(\mathbf{0}, \mathbf{I}),
\end{gather}    
where $\bm{\alpha}:=(\alpha_1,\ldots,\alpha_T)\in\mathbb{R}_{\ge0}^T$ controls the noise schedule. The reverse process then reconstructs $\mathbf{x}_0$ from $\mathbf{x}_T$. This paradigm now underpins a broad spectrum of visual synthesis and restoration tasks, including image restoration and enhancement~\citep{liDiffusionModelsImage2025} and faithful image super-resolution~\citep{wangColoringNoiseAdversarial2026}.
In text-guided diffusion~\citep{takagiHighresolutionImageReconstruction2022}, conditioning on a text prompt $P$ is achieved via a text encoder (e.g., CLIP~\citep{radfordLearningTransferableVisual2021}). The denoiser $\epsilon_\phi$ is trained to predict the noise $\boldsymbol{\epsilon}$ given the noisy input $\mathbf{x}_t$ and text condition:
\begin{equation}
    L(\phi) = \mathbb{E}_{t\sim U(1, T), \bm{\epsilon} \sim\mathcal{N}(\mathbf{0}, \bm{I})} \| \bm{\epsilon} - \bm{\epsilon}_{\phi}(\mathbf{x}_t; t, P) \|^2_2,
\end{equation}
To enhance text-image alignment, Classifier-Free Guidance (CFG)~\citep{hoClassifierFreeDiffusionGuidance2021} interpolates between conditional and unconditional predictions:
\begin{equation}
    \tilde{\bm{\epsilon}}_{\phi}(\mathbf{x}_t; t, \!P,s) \!=\! \bm{\epsilon}_{\phi}(\mathbf{x}_t; t, \emptyset) \!+\! s \left( \bm{\epsilon}_{\phi}(\mathbf{x}_t; t, \!P) \!-\! \bm{\epsilon}_{\phi}(\mathbf{x}_t; t, \emptyset) \right), \label{eq:CFG}
\end{equation}
where $s$ is the guidance scale and $\emptyset$ denotes the null prompt. In practice, negative prompts $P^{-}$~\citep{duCompositionalVisualGeneration2020} are used instead of $\emptyset$ to avoid undesired content in the generated samples.

\subsection{Score Distillation Sampling}
SDS, first proposed in DreamFusion~\citep{pooleDreamFusionTextto3DUsing2022}, employs vision priors from pre-trained 2D diffusion models to supervise 3D model optimization, establishing itself as a foundational learning paradigm for 3D generation.
Given a 3D representation model parameterized by $\theta$ and a differentiable renderer $g(\theta, \mathbf{c})$ that generates a rendered image $\mathbf{x}$ from camera pose $\mathbf{c}$, SDS aligns the probability distribution of $\mathbf{x}$ with the diffusion model prior $p(\phi)$.
Specifically, SDS introduces a score function $\epsilon_\phi(\mathbf{x}_t; t, P)$, which predicts noise $\hat{\epsilon}$ for a noisy image $\mathbf{x}_t$ conditioned on text prompt $P$. The discrepancy between $\hat{\epsilon}$ and the injected Gaussian noise $\epsilon$ defines the gradient for updating $\theta$, thereby improving the 3D model.
The SDS gradient is computed as:
\begin{equation} \label{eq:SDS}
\nabla_{\theta} \mathcal{L}_{\rm{SDS}}(\phi, \mathbf{x} \!=\! g(\theta)) \triangleq \mathbb{E}_{t, \epsilon} \left[ w(t) \left( {\tilde{\epsilon}}_{\phi}(\mathbf{x}_t; t, P) \!-\! \epsilon \right) \frac{\partial \mathbf{x}}{\partial \theta} \right],
\end{equation} 
where $w(t)$ is a time-dependent weighting function.

\begin{figure*}[t]
  \centering
  \includegraphics[width=.95\textwidth]{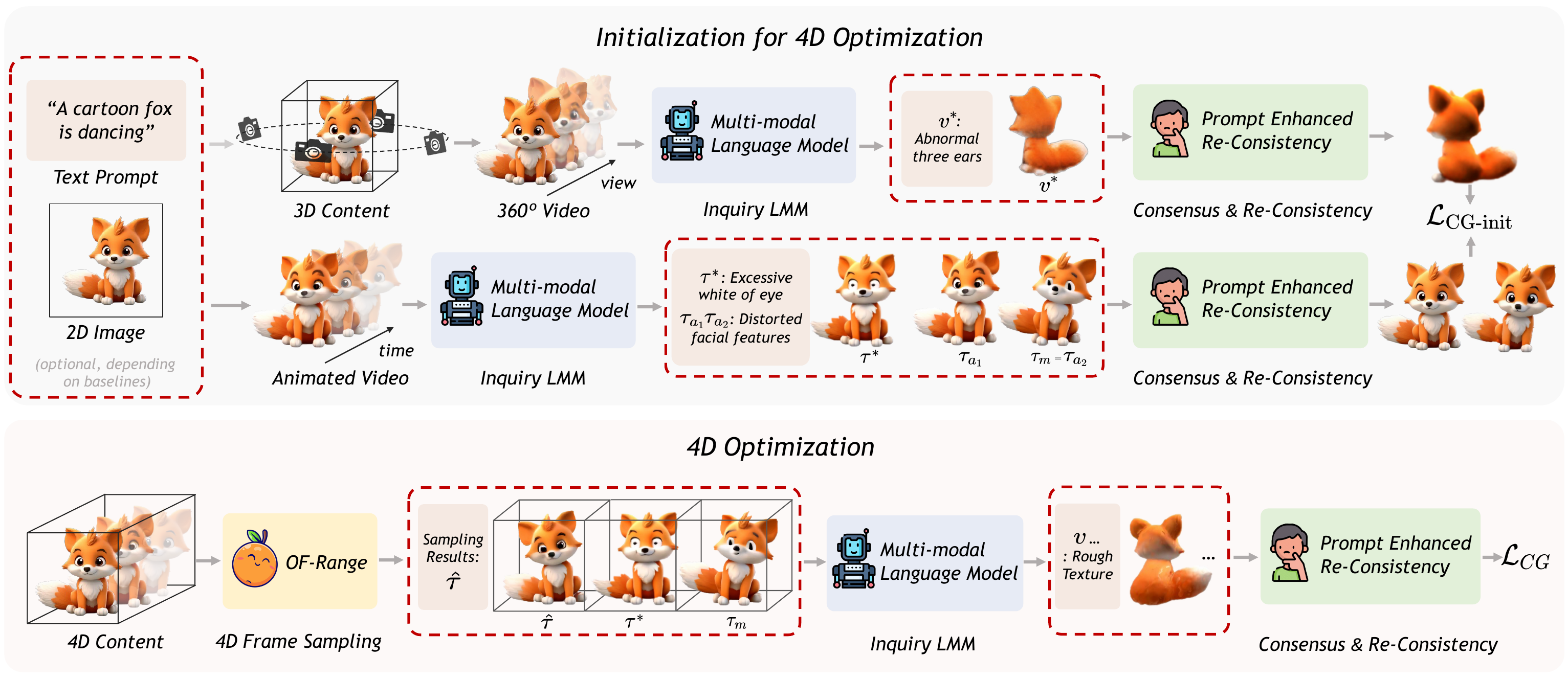}
  \caption{\textbf{Pipeline overview of the 4D spatiotemporal consistency stage.} We first identify the views and frames with the most severe inconsistencies through initialization to guide early-stage training. During training, we employ \textit{OF-Range}, an optical flow-based inter-frame sampling strategy that improves both the efficiency and effectiveness of the 4D generation process.
}
  \label{fig:pipeline_4d}
\end{figure*}

\section{Methodology}
This section introduces the overall framework and underlying principles of Hallo4D. At the core of Hallo4D lies a consistency optimization strategy for 3D content generation, which serves as the foundation for addressing a more challenging aspect—achieving spatiotemporal consistency in the 4D domain. Building on this foundation, we first describe our method for enhancing multi-view consistency in 3D generation. We then extend this approach to incorporate temporal coherence, enabling joint optimization across both spatial and temporal dimensions.

\subsection{Multi-view Appearance Alignment} \label{sec:3.2_multi-view}
Conventional SDS optimizes isolated single-view images, which contradicts the principle of multi-view simultaneity required for consistent 3D representation. This single-view formulation limits cross-view interaction during training, leading to the loss of surface details during denoising, as shown in Fig.\ref{fig:ablation}. To address this, we propose a \textit{Multi-view Appearance Alignment} strategy by introducing an appearance-attention mechanism $\mathcal{AA}{ttn}(\cdot)$ into the denoising function $\tilde{\epsilon}_\phi(\cdot)$, replacing the original self-attention in the U-Net architecture\citep{ronnebergerUNetConvolutionalNetworks2015}.

Specifically, inspired by recent advancements in diffusion models \citep{liuZePoZeroShotPortrait2024a, caoMasaCtrlTuningFreeMutual2023}, which suggest that query features within the self-attention spaces primarily shape image structure and layout, while key and value features influence texture, our method leverages this insight. As illustrated in the bottom left corner of Fig.~\ref{fig:pipeline_3d}, we select a focal view $i$ based on the camera pose, using the image from this viewpoint to provide the key and value features in the attention module. These are used to compute query features across all views, ensuring alignment of appearances. The cross attention is defined by the following formula:
\begin{equation} \label{eq:aattn}
\mathcal{AA}ttn(Q,K_i,V_i) = \text{Softmax}\left(\frac{QK_{i}^T}{\sqrt{d}}\right) \cdot V_{i},
\end{equation}
where $\mathcal{AA}ttn(\cdot)$ is the appearance attention, with \( K_i \) and \( V_i \) as the key and value features corresponding to the image rendered from the focal view \( i \), and \( Q \) as the query feature from all views. The key and value are derived from the focal view, while each of the four views computes a distinct query. By jointly rendering and processing images from multiple viewpoints, our method enables view-aware denoising and enhances appearance consistency across views, outperforming conventional single-view SDS approaches~\citep{pooleDreamFusionTextto3DUsing2022, wangScoreJacobianChaining2022}

\subsection{Multi-modal Hallucination Detection} \label{sec:3.3_LMM}
Existing 3D generation methods often suffer from spatial inconsistencies, primarily due to the inherent limitations of 2D pre-trained models in understanding 3D geometry. The problem become even more pronounced in 4D generation, where temporal dynamics further amplify spatial artifacts.
Moreover, while 3D generation benefits from the relatively standardized SDS framework~\citep{pooleDreamFusionTextto3DUsing2022}, 4D generation lacks a unified paradigm. Its diverse conditioning inputs—ranging from text, images, and videos to 3D assets—combined with the higher-dimensional nature of the task, have led to fragmented and less generalizable solutions.
This lack of coherence across methods impedes direct applicability to existing frameworks, highlighting the need for a pluggable and broadly compatible consistency optimization strategy.

\begin{figure*}[!t] 
    \centering
    \includegraphics[width=.95\linewidth]{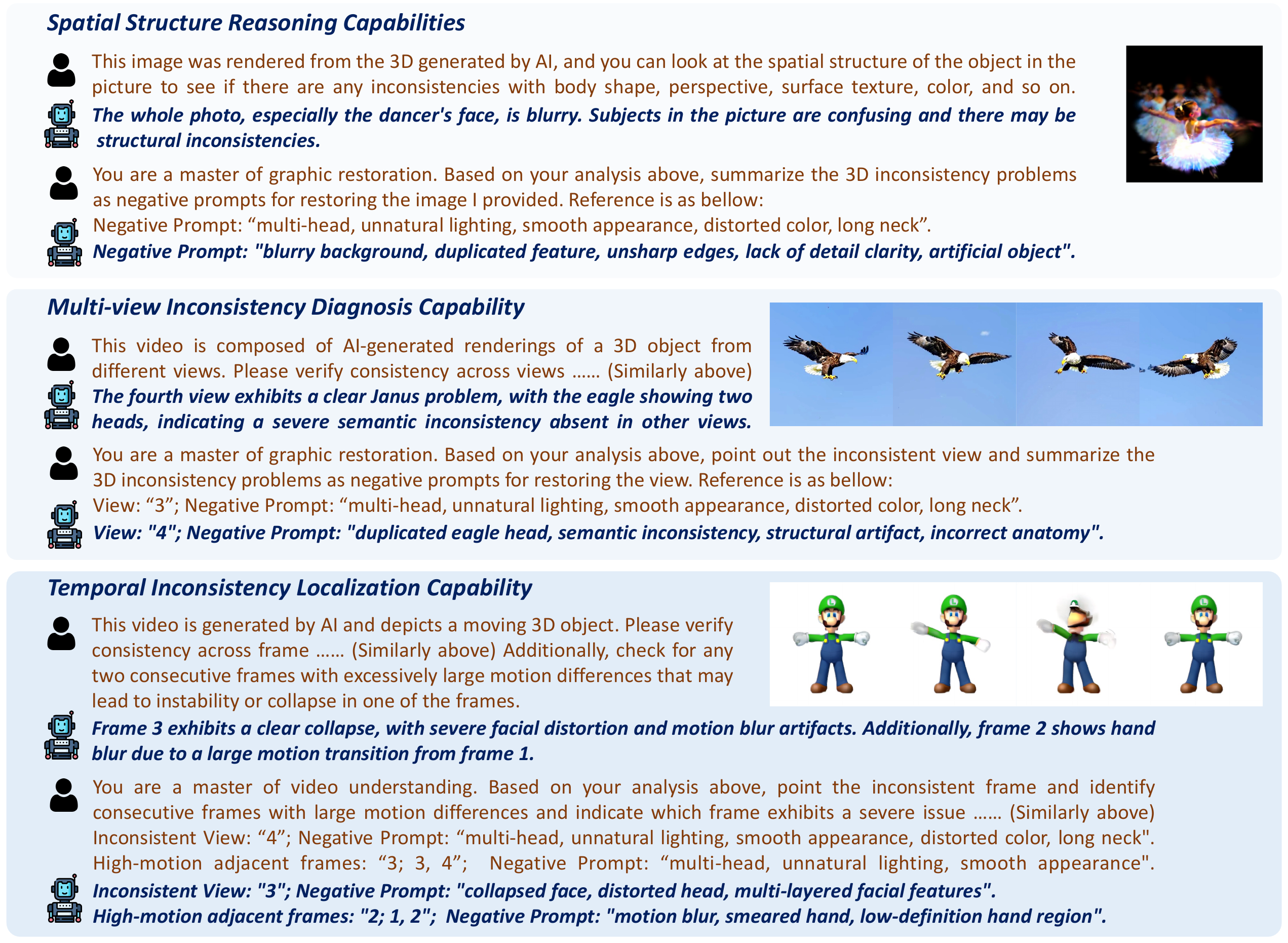}
    \caption{\textbf{A multi-modal reasoning case study} designed to evaluate the capabilities of LMMs in spatial structure reasoning, multi-view inconsistency diagnosis, and temporal inconsistency localization. The first round of dialogue demonstrates that LMMs possess the necessary reasoning abilities, while the second round illustrates that LMMs can produce responses in specific formats, enabling the subsequent extraction of both negative prompts and the associated target views or specific frames to be processed using regular expressions.}
    \label{fig:LMM}
\end{figure*}

To bridge this gap, we propose an inconsistency detection module, \textit{Multi-modal Hallucination Detection}, which leverages LMMs to analyze multi-view and multi-frame renderings—serving as the common basis for 4D generation. 
Benefiting from the strong cross-modal reasoning and temporal understanding capabilities of recent LMMs~\citep{liLLaVAOneVisionEasyVisual2024}, and in line with recent progress on multimodal hallucination evaluation and detection~\citep{chenSurveyMultimodalHallucination2026}, our approach builds on the hypothesis that LMMs can effectively detect and help mitigate spatiotemporal inconsistencies. 
To validate this hypothesis, we design a two-phase inquiry-based evaluation assessing three critical aspects: spatial structure reasoning, multi-view inconsistency diagnosis, and temporal inconsistency localization. As shown in Fig.~\ref{fig:LMM}, LMMs, even without explicit geometric supervision, can accurately identify inconsistencies across 3D renderings and video frames. 
Importantly, our use of LMMs is deliberately narrow: they act as lightweight view/frame-level consistency checkers in a constrained setting (single foreground subject, fixed-orbit cameras, standardized prompts), requiring pattern-level comparison rather than full 3D reconstruction. This aligns with their pretraining, which provides cross-view priors from multi-view/3D data. Concretely, LLaVA-OneVision-72B~\citep{liLLaVAOneVisionEasyVisual2024} is trained on multi-view/3D-related corpora, including~\citep{daiScanNetRichlyannotated3D2017, caesarNuScenesMultimodalDataset2020, shridharALFREDBenchmarkInterpreting2020, suhrCorpusReasoningNatural2019, jhamtaniLearningDescribeDifferences2018, hosseinzadehImageChangeCaptioning2021,azumaScanQA3DQuestion2022}, providing cross-view and 3D localization priors leveraged by our pipeline. To stabilize and standardize outputs, we adopt one-shot Chain-of-Thought prompting~\citep{brownLanguageModelsAre2020} with a reference template, from which we extract the indicated target views/frames and the corresponding negative prompts. 

In our framework, we consider three types of inputs: a single rendered image $\mathbf{x}^r$, a multi-view video $\mathbf{x}^v$ generated from 3D object renderings, and a temporally ordered front-view video $\mathbf{x}^f$, each paired with a 3D-aware inquiry prompt $P_I = \{P_{Ir}, P_{Iv}, P_{If}\}$.
The LMM processes these inputs to generate enhanced negative prompts $P_E^-$, which are then used to guide subsequent image-space corrections. The formulation is given by:
\begin{equation}
P^{-}_E =\mathcal{D}_{\psi}(\mathbf{x}^r,P_{Ir}),
\end{equation}
where $\mathcal{D}_\psi$ denotes the LMM parameterized by $\psi$. For 4D generation, the LMM additionally performs inconsistency reasoning over both views and frames:
\begin{align} \label{eq:init_4d}
   &v^*, P^{-}_E  =\mathcal{D}_{\psi}(\mathbf{x}^v,P_{Iv}), \\
   &\tau_m,\tau_{a1},\tau_{a2}, \tau^* ,P^{-}_E  =\mathcal{D}_{\psi}(\mathbf{x}^f,P_{If}),
\end{align}
where $v^*$ is the inconsistent view from multi-view renderings, $\tau^*$ is the detected inconsistent frame in the animated video, $\tau_{a1}$ and $\tau_{a2}$ are adjacent frames with significant motion differences, and $\tau_m \in \{\tau_{a1}, \tau_{a2}\}$ is the frame where inconsistency arises. For clarity, we omit explicit distinctions between different $P_E^-$ variants.

\subsection{Consensus-Driven Re-consistency via Prompt Enhancement} \label{sec:3.4_reconsistency}
With the enhanced negative prompt $P^{-}_E$ introduced in Sec.~\ref{sec:3.3_LMM}, one natural approach to improve consistency is to apply 2D image editing techniques directly on the rendered views.
However, a conceptual gap emerges: while image editing operates in the pixel space, most existing 3D generation pipelines grounded in SDS 
supervise 3D models through the noise space. This apparent mismatch raises concerns about the theoretical validity of applying image-space corrections under SDS-based training.
Recent works\citep{kingmaVariationalDiffusionModels, dengVariationalSchrodingerDiffusion2024} suggest that diffusion models can be reinterpreted to predict images directly rather than denoising noise. Motivated by this perspective, we revisit the underlying formulation of SDS to explore whether it can be consistently extended to the image space.

Starting from the DDPM, by substituting the expression of Eq.~(\ref{eq:forward_ddpm}) into Eq.~(\ref{eq:SDS}) to replace $\epsilon$, we obtain the following equation:
\begingroup
\small
\begin{equation} \label{eq:SDS_xspace}
\begin{aligned}
\nabla_{\!\theta} \mathcal{L}_{\rm{SDS}}(\phi,\mathbf{x}) \triangleq & \mathbb{E}_{t, \epsilon} \biggl[ w(t) \biggl( \hat{\epsilon}_{\phi}(\mathbf{x}_t; t, P) - \frac{\mathbf{x}_t \!-\! \sqrt{\alpha_t}\mathbf{x}_0}{\sqrt{1\!-\!\alpha_t}} \biggr) \frac{\partial \mathbf{x}}{\partial \theta} \biggr], \\
\triangleq & \mathbb{E}_{t, \epsilon} \biggl[ w^{\prime}(t) \Bigl( \mathbf{x}_0\!-\!\mathbf{x}_0^{\phi} \Bigr) \frac{\partial \mathbf{x}}{\partial \theta} \biggr],
\end{aligned}
\end{equation}
\endgroup
where 
\begin{equation}
    \mathbf{x}_0^{\phi} = \mathbf{x}_t \!-\! \frac{\sqrt{1 \!-\! \alpha_t} \hat{\epsilon}_{\phi}(\mathbf{x}_t; t, P)}{\sqrt{\alpha_t}}, \quad\!\!\!\!
    w^{\prime}(t) = \sqrt{\frac{\alpha_t}{1 \!-\! \alpha_t}} w(t),
\end{equation}
Building on this foundation, we derive an image-space formulation of SDS, thereby establishing the methodological validity of directly operating on rendered images. However, existing image editing approaches predominantly focus on manipulating the null prompt \citep{mokadyNulltextInversionEditing2023} or modifying the positive prompt \citep{hertzPrompttoPromptImageEditing2022}, which are generally ineffective at altering the geometric structure of 2D images rendered from 3D models. To overcome this limitation, we propose a novel module, termed Prompt-Enhanced Re-consistency, which introduces $P^{-}_E$ to enhance the geometric fidelity of the rendered outputs.

We regenerate the 2D rendered image \( \mathbf{x}_0 \) under the guidance of \( P^{-}_E \). Specifically, to preserve the original semantic information of \( \mathbf{x}_0 \), we employ DDIM \citep{songDenoisingDiffusionImplicit2022} to invert the image \( \mathbf{x}_0 \) to its noisy representation \( \mathbf{x}_T \) as follows.
\begin{align}
\hat{\mathbf{x}} _ {t-1} &= \sqrt{\frac{\alpha _ {t-1}} {\alpha _ t}} \hat{\mathbf{x}} _ t +  \delta_t
\tilde{\epsilon} _ {\phi} (\hat{\mathbf{x}} _ t; t, P, P^{-} _ E), \\
\delta_t &= (\sqrt{1-\alpha _ {t-1}}-\sqrt{\frac{\alpha _ {t-1} (1-\alpha _ t)} {\alpha _ t}}),
\end{align}
Subsequently, we apply DDIM sampling to regenerate the consistent versions of the image from the DDIM inverted \( \mathbf{x}_T \), denoted as \( \hat{\mathbf{x}}_0 \). This approach also ensures that the regenerated image retains its core semantic integrity while improving its multi-view consistency, and it can be readily derived by mathematical induction that:
\begin{align} \label{eq:x0_ddim}
    \hat{\mathbf{x}}_0 & =\sqrt{\frac{\alpha_0}{\alpha_T}}\mathbf{x}_T+\sum\limits_{i=1}^{T} \kappa_i
    \tilde{\epsilon} _ {\phi} (\hat{\mathbf{x}} _ t; t, P, P^{-} _ E), \\
    \kappa_i & = (\sqrt{\frac{\alpha_0(1-\alpha_{i-1})}{\alpha_{i-1}}}-\sqrt{\frac{\alpha_0(1-\alpha_i)}{\alpha_i}}),
\end{align}

Meanwhile, a practical limitation of single-shot image editing is that it may introduce stochastic editing noise or local over-corrections, and repeated reliance on one edited output can accumulate error, especially when the editing model is imperfect. To reduce over-dependence on any single edit, we augment \textit{Prompt-Enhanced Re-consistency} with a consensus-based selection module inspired by the \textit{LMM Consensus Selector}. Instead of producing only one $\hat{x}_0$, we generate a candidate set of edited images by varying the editing randomness which includes DDIM sampling noise and timesteps:
\begin{equation}
\mathcal{X}_{\text{cand}} = \{\hat{\mathbf{x}}_0^{(1)}, \hat{\mathbf{x}}_0^{(2)}, \ldots, \hat{\mathbf{x}}_0^{(n)}\},
\end{equation}

We then introduce an evaluator ensemble of LMMs $\mathcal{D}^{\text{eval}}_{\psi}=\{\mathcal{D}^{\text{eval}}_{\psi1}, \ldots, \mathcal{D}^{\text{eval}}_{\psi m}\}$, each independently scoring the candidates with respect to 
(i) structural integrity, (ii) prompt alignment, and (iii) cross-view/temporal plausibility under the same constrained setting used by our detector. Each evaluator selects the index of its preferred candidate:
\begin{equation}
\gamma_k = \arg\max_{\gamma} \,S_k^{\left(\gamma\right)}, 
\quad 
S_k^{\left(\gamma\right)} = \mathcal{D}^{\text{eval}}_{\psi k}\!\left(\hat{\mathbf{x}}_0^{(\gamma)}, P_S\right)
\end{equation}
where $k \in [m]$, $S_k^{(\gamma)}$ denotes the score assigned by the $k$-th LMM to the $\gamma$-th image, and $P_S$ represents the scoring prompt provided to the LMMs. Then we aggregate votes across evaluators:
\begin{equation}
\text{Vote}(\hat{\mathbf{x}}_0^{(\gamma)}) = \sum_{k=1}^{m} \mathbb{I}[\gamma_k = \gamma],
\end{equation}
The final corrected image is chosen by majority voting, 
\begin{equation}
\hat{\mathbf{x}}_0^{*}=\arg\max_{\gamma}\mathrm{Vote}(\hat{\mathbf{x}}_0^{(\gamma)}),
\end{equation}
This consensus mechanism explicitly targets the failure mode where a single edit introduces subtle artifacts that are hard to detect numerically but are consistently penalized by multiple independent LMM judges, thereby improving robustness and reducing the risk of error accumulation across optimization iterations.

\begin{figure}[t]
    \centering
    \includegraphics[width=0.97\linewidth]{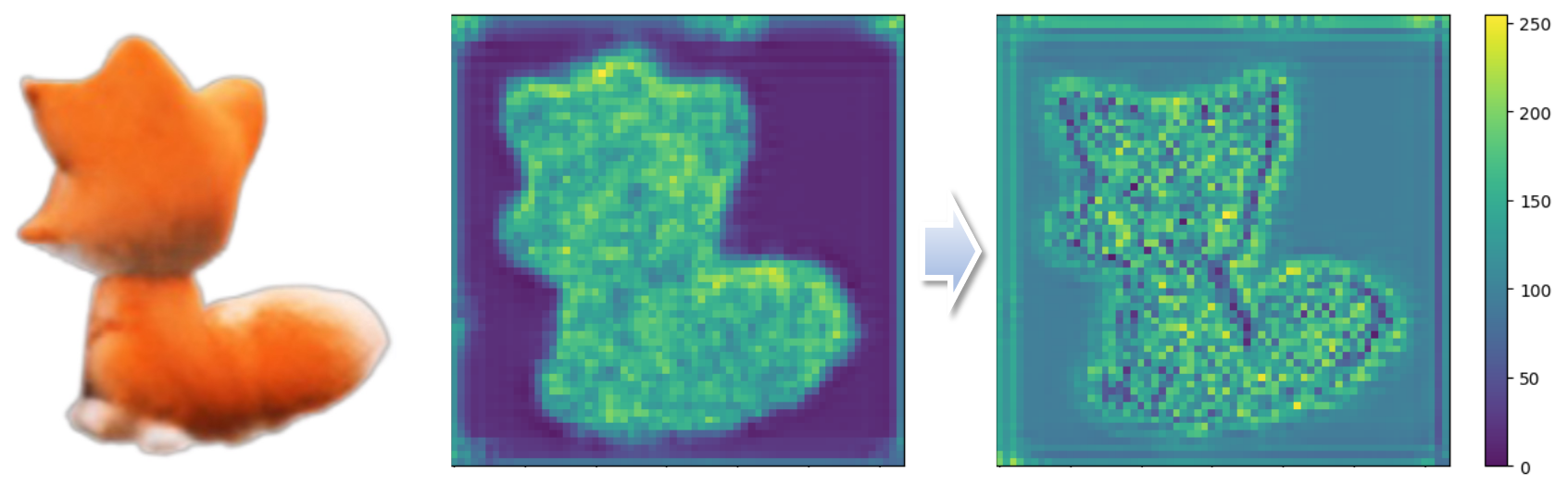}
    \caption{\textbf{Attention Visualization in Prompt-Enhanced Re-consistency}. \textit{Left:} image shows an inconsistent rendered view. \textit{Middle and Right:} images visualize U-Net attention from the DDIM-Inversion-based image editing model, with the middle and rightmost images corresponding to early ( $t \rightarrow T$ ) and late ( $t \rightarrow 0$ ) diffusion steps, respectively. It is evident that, as diffusion progresses, the attention gradually decreases its focus on the inconsistent regions.}
    \label{fig:attention}
\end{figure}

Finally, we train the 3D model $\theta$ using the MSE loss \(\mathcal{L}_{\rm{CG}}\) between \(\mathbf{x}_0\) and \(\hat{\mathbf{x}}_0\) in the image space:
\begin{equation} \label{eq:loss_cg}
\mathcal{L}_{\rm{CG}} \triangleq \mathbb{E} \left[\left| \hat{\mathbf{x}}_0^* - \mathbf{x}_0 \right|^2 \right],
\end{equation}
By combining Eq.~(\ref{eq:SDS_xspace}) and Eq.~(\ref{eq:loss_cg}), the effectiveness of our method can be theoretically justified, provided that $\hat{\mathbf{x}}_0$ exhibits stronger view consistency and better alignment with realistic semantics than $\mathbf{x}_0^{\phi}$. To support this, we visualize the attention maps from the intermediate steps of Eq.~(\ref{eq:x0_ddim}), as shown in Fig.~\ref{fig:attention}. It can be seen that during the early stages of editing, the attention covers the entire fox in the image, while in the later stages, attention to inconsistent regions—such as the third ear on top of the fox’s head—is significantly reduced. This indicates that the negative promot-based image editing technique effectively corrects inconsistencies in the rendered image, producing a refined $\hat{\mathbf{x}}_0$ that better aligns with realistic semantics.
The overall process is illustrated in Fig.~\ref{fig:pipeline_3d}.

Moreover, it is worth noting that we apply Prompt-Enhanced Reconsistency only when the rendered image exhibit complete semantic structure. Our detector, \( \mathcal{D}_{\psi} \), assesses the semantic completeness of the image. If the semantic structure is deemed incomplete or unclear, \( \mathcal{D}_{\psi} \) returns None, precluding further processing. This ensures that enhancements are only applied to images that are adequately prepared. 
This dependency of our enhancement process on the state of semantic completeness directly influences the formulation of the final training loss for the 3D model, as detailed below:
\begin{equation} \label{eq:loss}
    \mathcal{L}(\theta)=\left\{
    \begin{aligned}
    &\mathcal{L}_{\rm{SDS}} + w_{1}\mathcal{L}_{\rm{CG}}, &&\text{if } \mathcal{D}_{\psi}(\mathbf{x},P_I) \ne\text{None},\\
    &\mathcal{L}_{\rm{SDS}}, &&\text{otherwise}, 
    \end{aligned}\right.
\end{equation} 
where $w_1$ is set to balance the magnitude of $\mathcal{L}_{\rm{SDS}}$ and $\mathcal{L}_{\rm{CG}}$. By incorporating $\mathcal{L}_{CG}$, which is only applied when \( \mathcal{D}_{\psi} \) confirms the semantic readiness of the image, we ensure that our model focuses on enhancing well-formed images. This selective application of $\mathcal{L}_{CG}$ prevents further exacerbating the quality of images already of poor quality. Simultaneously, it avoids misallocating resources to images that do not benefit from the intended enhancements, thereby improving the efficiency and effectiveness of our training process.

\subsection{Initialization for 4D Optimization} \label{sec:3.5_init4d}

As discussed in Sec.~\ref{sec:3.3_LMM}, 4D generation poses more severe spatiotemporal consistency challenges compared to 3D counterpart. Through our experiments, we further observe that the diverse input modalities in 4D generation—whether from the 3D object or the animated video—can independently introduce inconsistencies that propagate and degrade the final output. This highlights the necessity of early intervention during model training. To address this, we propose an initialization strategy for 4D optimization that preprocesses inconsistencies in both the 3D content and the animated video, thereby mitigating error accumulation during the subsequent 4D training process.

\begin{figure}[t]
    \centering
    \includegraphics[width=0.98\linewidth]{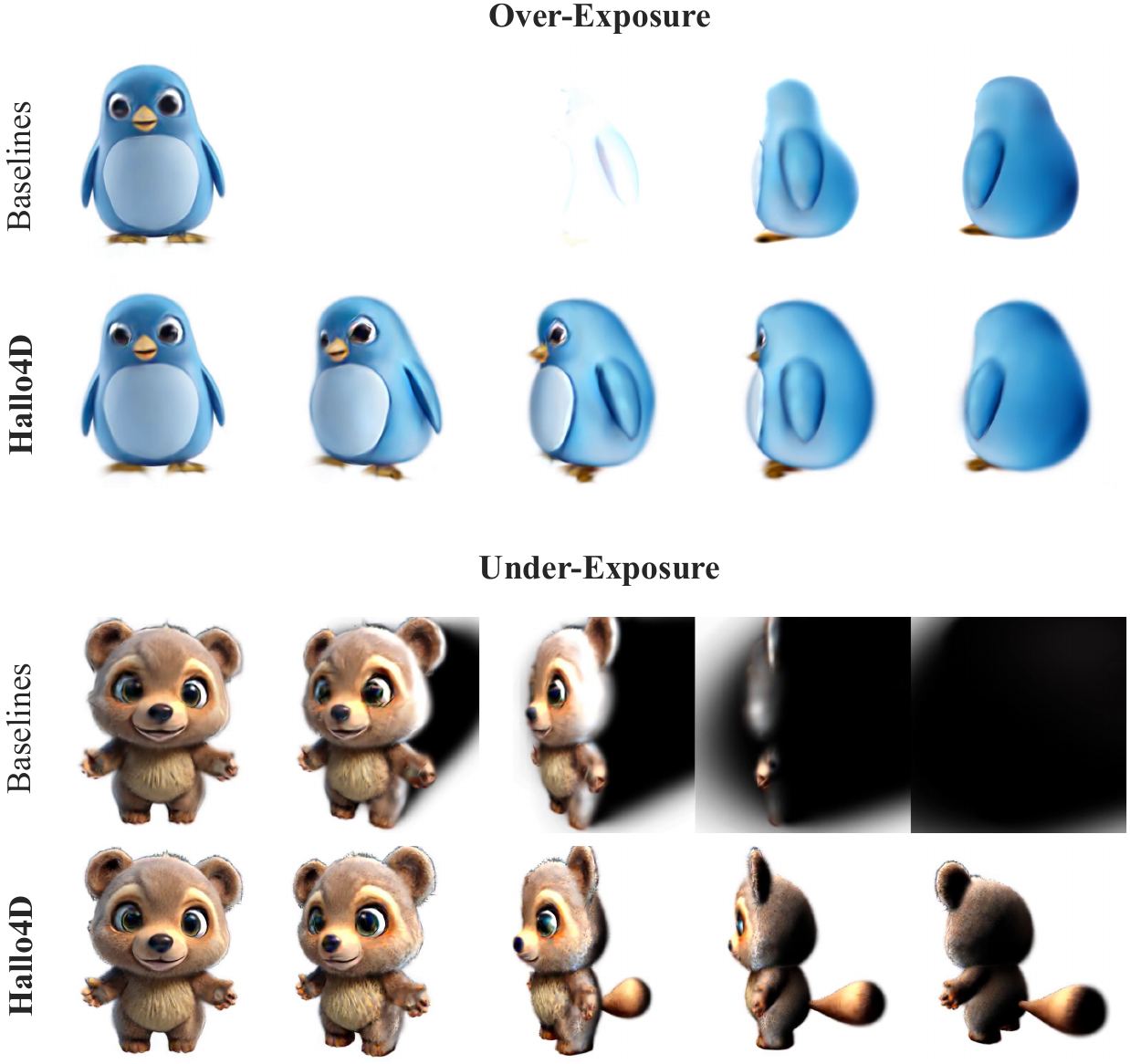}
    \caption{\textbf{Illustration of over-exposure and under-exposure situation.} Baselines exhibit view-dependent collapse in non-frontal views, producing white-out/black-out frames, whereas Hallo4D, with losses of \textit{CSEA} and \textit{LDR}, maintains stable exposure and preserves structure across viewpoints.}
    \label{fig:exposure}
\end{figure}

Specifically, for the 3D object in the 4D generation input, we sample eight surrounding views at 45° intervals and concatenate them in order to form a video. This video is then fed into the LMM, which is queried to identify the view exhibiting inconsistency and to summarize the issue as a negative prompt, yielding $v^*, P_E^-$. For the animated video, it is directly input to the LMM, which is asked to identify the inconsistent frame $\tau^*$, the pair of adjacent frames with significant motion $\tau_{a1}, \tau_{a2}$, and the specific frame $\tau_{m}$ within the pair where the inconsistency occurs due to large motion. These inconsistency cues are then summarized into the corresponding $P_E^-$. The process is formalized in Eq.~(\ref{eq:init_4d}) and illustrated in Fig.~\ref{fig:pipeline_4d}.
In addition, we apply the Prompt-Enhanced Re-Consistency module to the identified inconsistent views or frames, and following Eq.~(\ref{eq:loss_cg}) compute the corresponding loss term $\mathcal{L}_{\rm{CG}-init}$, which is weighted and applied during the first 10\% of training iterations on top of the baseline. The formulation is as follows:
\begin{equation} \label{eq:loss_4d}
    \mathcal{L}_{\text{4D}}(\theta)=\left\{
    \begin{aligned}
    &\mathcal{L}(\theta) + w_{2}\mathcal{L}_{\rm{CG}-init}, &&\text{if } e<10\%\rm{E}, \\
    &\mathcal{L}(\theta), &&\text{otherwise}, 
    \end{aligned}\right.
\end{equation} 
where $w_2$ is a weighting factor that balances the scale of different loss terms, $e$ denotes the current training epoch, and $\rm{E}$ represents the total number of training epochs.

\begin{figure*}[!t]
  \centering
  \includegraphics[width=0.94\textwidth]{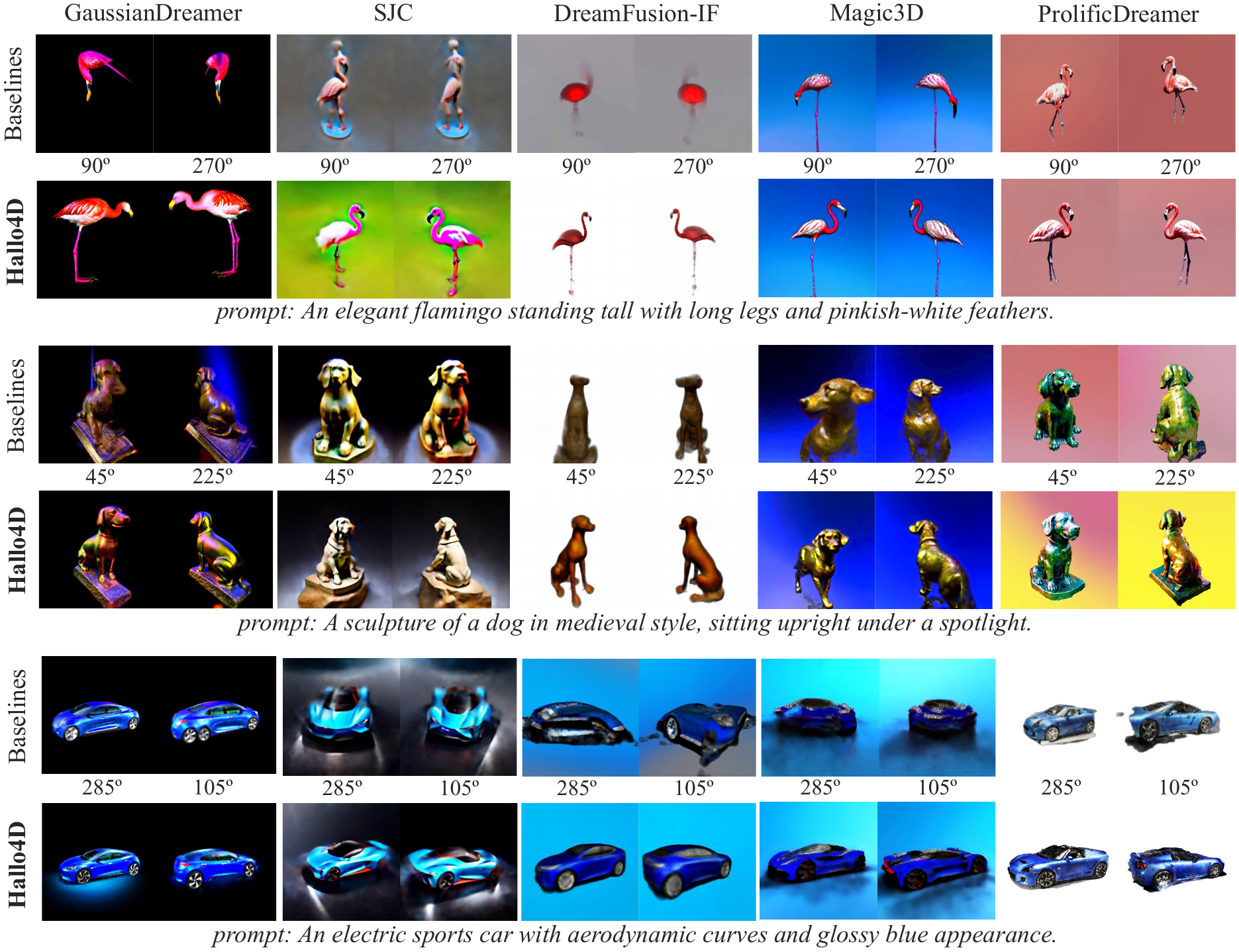}
  \caption{\textbf{Qualitative comparison in text-driven 3D generation of Hallo4D and baseline models.} For intuitive comparison, we use the same and complementary viewpoints to present the results, where our method shows significant improvements.}
  \label{fig:expr_text_to_3D}
\end{figure*}

\subsection{OF-Range based Sampling Strategies} \label{sec:3.6_ofrange}
Although multi-modal hallucination detection via LMMs enables accurate identification of spatiotemporal inconsistencies, querying LMMs incurs non-negligible computational overhead, making full-frame or full-view detection impractical at every training epoch.
In 3D generation, we alleviate this by randomly selecting one view per epoch. However, this strategy is unreliable for 4D keyframe selection: as noted in Sec.~\ref{sec:3.5_init4d}, the known frames $\tau_m$ and $\tau^*$ are easily oversampled, reducing optimization efficiency.
Meanwhile, extensive research\citep{laiLearningBlindVideo2018, huangVBenchComprehensiveBenchmark2023, yangFrescoSpatialTemporalCorrespondence2024a} underscores the importance of effective frame selection for maintaining temporal consistency.
A natural solution is to detect keyframes based on geometric changes across time. However, synthesized 3D objects often exhibit free-form motion without rigid-body assumptions, and localized deformations cannot be reliably captured by explicit geometric features.

Therefore, we propose \textit{OF-Range}, an optical flow-based inter-frame sampling strategy that improves both the efficiency and effectiveness of the 4D generation process.
Specifically, we begin by rendering a sequence of front-view frames from the 4D asset, resulting in a frame set. For each frame $\tau$, we extract Shi-Tomasi corner \citep{jianboshiGoodFeaturesTrack1994} features $C_{\tau} = \{c_j\}$ from geometrically stable interior regions, leveraging cross-frame coherence while avoiding transient surfaces that may introduce reconstruction ambiguity. Optical flow vectors $\{\mathbf{v}_j\}$ between consecutive frames are then computed using the Lucas-Kanade method\citep{lucasIterativeImageRegistration1981}. These are used to define a deformation-sensitive motion saliency score:
\begin{equation}
\mathcal{S}_{\tau} = \frac{\sum_j \exp(-\lambda d(c_j, \Gamma_{\tau})) |\mathbf{v}_j|}{\sum_j \exp(-\lambda d(c_j, \Gamma_{\tau}))},
\end{equation}
where $\exp(-\lambda d(c_j, \Gamma_{\tau}))$ is a spatial confidence weight that decays exponentially with the Euclidean distance $d(c_j, \Gamma_{\tau})$ from each feature point $c_j$ to the nearest unstable boundary region $\Gamma_{\tau}$. This weighting mechanism suppresses noisy motion near volatile areas while amplifying deformation cues from stable regions.

Given that most 4D generation models operate on short sequences (typically $n < 15$) due to GPU memory constraints, it is crucial to prevent selection bias and avoid repeatedly sampling frames near the previously identified keyframes $\tau_m$ and $\tau^*$. To this end, we formulate keyframe selection as a regularized stochastic optimization problem:
\begin{gather} \label{eq:ofrange_tau}
\hat{\tau} = \mathop{\arg\max}_{\tau} \{\mathbb{E}[\mathcal{S}_{\tau}] - \gamma \bigl(\mathbb{I}(\tau_m \neq \tau^*) \cdot Q_{\tau} \bigl) + G_{\tau}\}, \\
Q_{\tau} = \min(|\tau - \tau_m|, |\tau - \tau^*|)/n,
\end{gather} 
where $Q_{\tau}$ penalizes the selection of frames temporally close to existing keyframes, and $G_{\tau} \sim \text{Gumbel}(0, 1)$ introduces stochasticity to encourage diverse sampling across epochs. This strategy jointly accounts for motion saliency, temporal diversity, and stochastic exploration, leading to more robust and efficient keyframe selection for 4D training.

\subsection{Adaptive Exposure-Aware Semantic Alignment}
Hard cases remain a persistent obstacle in 4D generation, with one of the most severe failure modes being view-dependent collapse under non-frontal viewpoints. In our empirical studies, we repeatedly observe that exposure instability—either as a primary error or via propagation through the generation pipeline—can precipitate catastrophic overexposure or underexposure, yielding outputs that are nearly uniform extreme white or black. In such cases, the underlying semantics are largely lost, where recovery is challenging for inversion-based post-hoc procedures. These exposure-driven failure modes are illustrated in Fig.~\ref{fig:exposure}.

To address these hard cases, we augment our existing strategy with a contrast-aware reward specifically aimed at stabilizing exposure in challenging views. Because background handling varies across baselines, we first factor out background effects by deriving a soft foreground prior from the alpha channel, $M\in[0,1]^{H\times W}$, and compositing a foreground-only image $\mathbf{x}^{\text{fg}}$ onto a neutral canvas $C$:
\begin{equation}
\mathbf{x}^{\text{fg}}=M\odot\mathbf{x}+(1-M)\odot C,
\end{equation}
This concentrates optimization on the object—routing gradients primarily through foreground pixels—while the neutral canvas suppresses spurious background influence. To combine signals from multiple exposure-re  lated loss components in a stable and differentiable way, we define the \textit{log-sum-exp} operator $\text{LSE}$ with parameter $\mu$ as:
\begin{equation}
\text{LSE}(\mathcal{Z};\mu)=\mu\log\sum_{z\in \mathcal{Z}}\exp({z/\mu}),
\end{equation}
This operator behaves as a smooth maximum: for small $\mu$ it approximates the worst offender in $\mathcal{Z}$, while larger $\mu$ yields a softer aggregation. It thus provides a tunable mechanism to focus the penalty on the most problematic exposure tendency without introducing non-differentiabilities. Concretely, a CLIP-based reward $\mathcal{R}c$ is computed on the foreground-only image $\mathbf{x}^{\text{fg}}$ using exposure-aware textual prompts. 
\begin{equation}
    \mathcal{R}_c^{P}=\mathrm{CLIP}(P,\mathbf{x}^{\text{fg}}|P\in P_{\Lambda}),
\end{equation}
where $P_{\Lambda}=\{P_{\Lambda}^{+},P_{\Lambda}^{\text{over}},P_{\Lambda}^{\text{under}}\}$ denote the well-exposed, over-exposed, and under-exposed prompts respectively, and define $\mathcal{R}_c^{P}=\mathrm{CLIP}(P,\mathbf{x}^{\text{fg}})$ as the corresponding CLIP-Score. These rewards serve as the inputs to the \textit{log-sum-exp} operator, which provides a smooth, stable means of combining evidence across exposure modes. The resulting \textit{Contrastive Semantic Exposure Alignment} loss encourages strong alignment with well-exposed semantics while simultaneously suppressing alignment with over- and under-exposed semantics, yielding a single differentiable objective that focuses learning on the most problematic exposure tendency.
\begin{equation}
\mathcal{L}_{\text{CSEA}}=-\mathcal{R}c^++\text{LSE}(\mathcal{R}c^{\text{over}},\mathcal{R}c^{\text{under}};\mu),
\end{equation}
We optimize a contrastive, exposure-aware objective that pulls the image toward the well-exposed description while pushing it away from both over- and under-exposed cues.
During optimization, gradients are masked to the foreground so that background brightness does not become a lever for objective reduction:
\begin{equation}
\nabla_{\mathbf{x}}\mathcal{L}_{\text{CSEA}}=M\odot ({\partial\mathcal{L}_{\text{CSEA}}}/{\partial\mathbf{x}^{\text{fg}}}),
\end{equation}
This gating eliminates background-driven degeneracies, stabilizes updates on the object surface, and ensures that improvements in the loss correspond to genuine object-level corrections rather than exposure tricks.

\begin{figure*}[!t]
  \centering
  \includegraphics[width=0.93\textwidth]{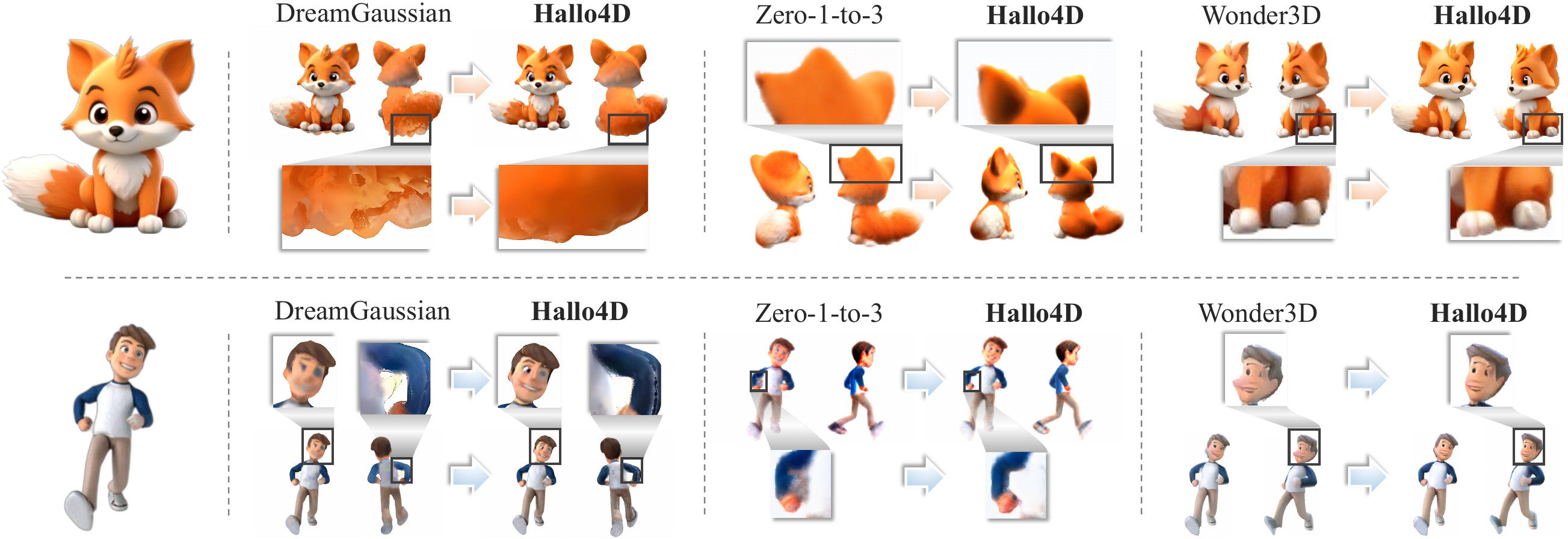}
  \caption{\textbf{Qualitative comparison in image-driven 3D generation of Hallo4D and baseline models.} For intuitive comparison, we enlarge the regions with noticeable differences, and the results demonstrate that our method achieves significant improvements.}
  \label{fig:expr_image_to_3D}
\end{figure*}

Moreover, we regularize the rendered appearance toward a target contrast bandwidth with a
\emph{log-dynamic-range} loss:
\begin{equation}
    \mathcal{L}_{\text{LDR}}=|\,\text{std}\,\bigl(\log(\vartheta+Y)\bigl)-\sigma_0|,
\end{equation}
where $\operatorname{std}(\cdot)$ is the per-image standard deviation evaluated over foreground pixels, $Y$ denotes per-pixel luminance in the linear RGB domain, $\vartheta\!>\!0$ is a small stabilizer that prevents singular gradients in dark regions, and $\sigma_{0}$ specifies the desired log-luminance standard deviation. Minimizing $\mathcal{L}_{\text{LDR}}$ steers the contrast distribution toward the desired bandwidth, suppressing both insufficient contrast and excessive contrast that would otherwise push highlights or shadows into saturation.

Across extensive experiments, we found that there exist cases where apparent under-exposure is in fact pseudo under-exposure: large swaths of pure-black floating geometry arise within the frustum in non-frontal views and occlude the subject. In parallel, we frequently observe floating clutter outside the camera frusta. For the out-of-frustum case, existing 3D/4D pipelines leave these regions unconstrained—such primitives are never rendered, receive no SDS supervision, and remain invisible until model export. To reduce the risk of unpredictable generation collapses and make effective use of limited memory, at each validation pass we apply deterministic visibility pruning that retains only primitives whose centers lie within the \textit{union of camera frusta (UoF)} defined by the evaluation trajectory and discards the rest. 
Although introduced to suppress out-of-frustum clutter, this simple UoF pruning also alleviates the aforementioned pseudo under-exposure by removing the pure-black floating geometry that occludes non-frontal views, thereby improving the robustness of the 4D generation pipeline.

\begin{figure*}[!t]
  \centering
  \includegraphics[width=0.96\textwidth]{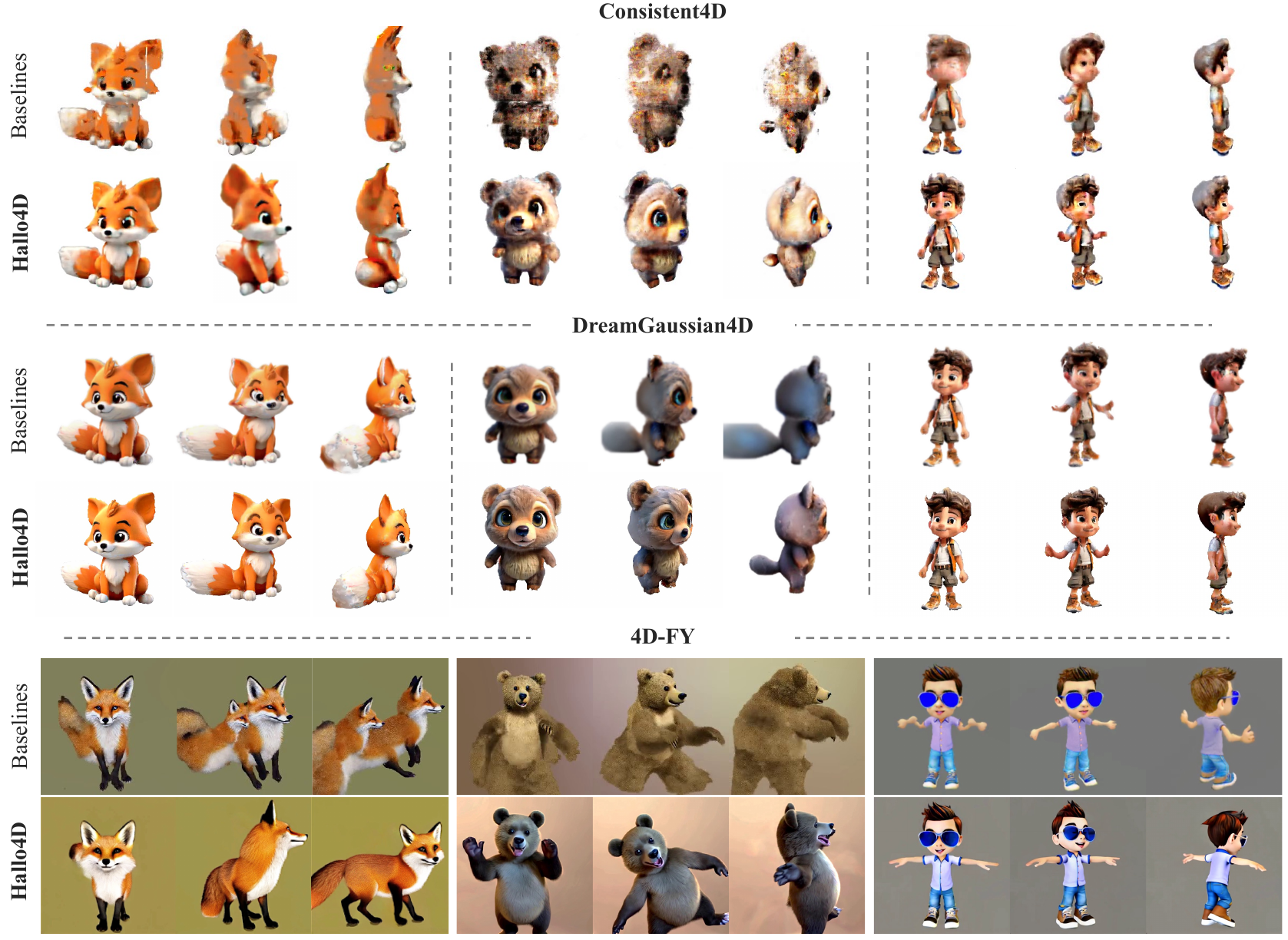}
  \caption{\textbf{Qualitative comparison in 4D generation of Hallo4D and baseline models.}  Among them, Consistent4D and DreamGaussian4D are image-to-4D methods, while 4D-FY is a text-to-4D generation approach.}
  \label{fig:expr_4D}
\end{figure*}

\section{Experiments}

\begin{table}[!t]
\centering
\caption{Quantitative comparisons in text-driven 3D generation.}
\renewcommand{\arraystretch}{1.3}
\begin{tabular}{l ccc ccc}
\toprule
\multirow{2}{*}{\textbf{Methods}} & \multicolumn{3}{c}{\textbf{CLIP Scores} $\uparrow$} & \multicolumn{3}{c}{\textbf{User Study $\uparrow$}}\\
\cmidrule(lr){2-4} \cmidrule(lr){5-7}
 & \textbf{B/32} & \textbf{B/16} & \textbf{L/14} & \textbf{Cons.} & \textbf{Qual.} & \textbf{Align.} \\
\midrule
GSD & 21.31 & 22.67 & 23.70 & 6.00 & 5.53 & 5.57 \\
\rowcolor{blue!7}\textbf{+Hallo4D} & \textbf{24.68} & \textbf{27.11} & \textbf{30.22} & \textbf{9.01} & \textbf{8.82} & \textbf{8.95} \\
SJC & 20.13 & 21.36 & 23.95 & 4.53 & 4.77 & 5.63 \\
\rowcolor{blue!7}\textbf{+Hallo4D} & \textbf{24.39} & \textbf{26.41} & \textbf{28.17} & \textbf{7.79} & \textbf{7.92} & \textbf{7.53} \\
DF-IF & 14.09 & 15.98 & 18.19 & 4.63 & 4.17 & 4.70 \\
\rowcolor{blue!7}\textbf{+Hallo4D} & \textbf{22.19} & \textbf{23.82} & \textbf{26.80} & \textbf{6.41} & \textbf{7.50} & \textbf{7.12} \\
Magic3D & 14.93 & 16.41 & 17.99 & 5.13 & 4.60 & 5.17 \\
\rowcolor{blue!7}\textbf{+Hallo4D} & \textbf{22.14} & \textbf{24.35} & \textbf{27.83} & \textbf{7.61} & \textbf{8.06} & \textbf{7.40} \\
P.Dreamer & 22.45 & 24.18 & 26.85 & 4.17 & 8.50 & 7.26 \\ 
\rowcolor{blue!7}\textbf{+Hallo4D} & \textbf{25.71} & \textbf{28.42} & \textbf{31.49} & \textbf{7.64} & \textbf{9.37} & \textbf{9.23} \\ 
\bottomrule
\end{tabular}
\label{tab:text-3d}
\end{table}

In this section, we conduct a comprehensive evaluation of Hallo4D on both 3D and 4D generation tasks, presenting comparative results against baseline models to showcase its effectiveness. To further demonstrate Hallo4D’s ability to enhance multi-view consistency in 3D and spatiotemporal consistency in 4D generation, we also perform an extensive user study. Finally, we carry out ablation experiments to validate the necessity of each component in the proposed framework.

\subsection{Experiment Setup}

\textbf{Implementation.} We set $w_1$ in Eq.~(\ref{eq:loss}) to $0.1$, and decay $w_2$ in Eq.~(\ref{eq:loss_4d}) from $0.1$ to $0.01$ over training. Additionally, due to the presence of $\mathcal{L}_{\text{CG-init}}$, we apply a decaying weight to $R_{\tau}$ in Eq.~(\ref{eq:ofrange_tau}), reducing it from $1$ to $0.0001$. To select the focal view in Fig.~\ref{eq:aattn}, we use the camera’s vertical field of view (Fovy), choosing the first view where Fovy exceeds 120\% of the baseline default.
For LMM reasoning, we employ a streamlined setup using a single interaction for efficiency—merging the two-stage dialogue in Fig.~\ref{fig:LMM} into one. To align with the general negative prompts used in existing baselines~\citep{yiGaussianDreamerFastGeneration2023, wangScoreJacobianChaining2022, pooleDreamFusionTextto3DUsing2022, tangDreamGaussianGenerativeGaussian2023, linMagic3DHighResolutionTextto3D2023, liuZero1to3ZeroshotOne2023}, we adopt:\textit{ “unnatural colors, poor lighting, low quality, artifacts, smooth texture.”}
For the MLLM Consensus Selector, we generate four candidate images per pass and employ an ensemble of four LMMs to score them.
For exposure adjustment, we set the LSE temperature to $\mu=0.06$. The prompts in $P_{\Lambda}$ are tailored by LMM prior to optimization. We compute luminance in the linear RGB domain as $Y=0.2126R+0.7152G+0.0722B$ and use $\vartheta=0.01$ and $\sigma_0=0.9$, chosen empirically.
We use LLaVA-OneVision-72B~\citep{liLLaVAOneVisionEasyVisual2024} as the LMM backbone for this task. All experiments are conducted on NVIDIA A100 GPU.

\begin{table*}[t]
\centering
\caption{Quantitative comparisons in image-driven 3D generation.}
\renewcommand{\arraystretch}{1.4}
\begin{tabular}{l ccc ccccc ccc}
\toprule
\multirow{2}{*}{\textbf{Methods}}
& \multicolumn{3}{c}{\textbf{CLIP Scores} $\uparrow$}
& \multicolumn{5}{c}{\textbf{Objective Metrics}}
& \multicolumn{3}{c}{\textbf{User Study $\uparrow$}} \\
\cmidrule(lr){2-4} \cmidrule(lr){5-9} \cmidrule(lr){10-12}
& \textbf{B/32} & \textbf{B/16} & \textbf{L/14} 
& \textbf{CD $\downarrow$} & \textbf{Vol. IoU $\uparrow$} & \textbf{PSNR $\uparrow$} & \textbf{SSIM $\uparrow$} & \textbf{LPIPS $\downarrow$}
& \textbf{Cons.} & \textbf{Qual.} & \textbf{Align.} \\
\midrule
DreamGaussian
& 31.77 & 32.13 &  32.94
& 0.0185 & 0.5861 & 16.502 & 0.8543 & 0.2025
& 8.40 & 9.23 & 8.55 \\
\rowcolor{blue!7}\textbf{+Hallo4D}
& \textbf{32.14} & \textbf{35.03} & \textbf{35.68} 
& \textbf{0.0185} & \textbf{0.6116} & \textbf{16.530} & \textbf{0.8801} & \textbf{0.1734}
& \textbf{9.29} & \textbf{9.66} & \textbf{9.13} \\
Zero-1-to-3
& 24.45 & 25.17 & 26.01 
& 0.0370 & 0.4824 & 13.433 & 0.7210 & 0.3926
& 7.25 & 6.10 & 8.30 \\
\rowcolor{blue!7}\textbf{+Hallo4D}
& \textbf{25.13} & \textbf{25.71} & \textbf{26.89} 
& \textbf{0.0287} & \textbf{0.5614} & \textbf{14.939} & \textbf{0.7531} & \textbf{0.3335}
& \textbf{7.86} & \textbf{7.25} & \textbf{9.01} \\
Wonder3D
& 27.31 & 28.08 & 28.92 
& 0.2147 & 0.5218 & 12.258 & 0.7831 & 0.3823
& 8.69 & 8.21 & 8.76 \\
\rowcolor{blue!7}\textbf{+Hallo4D}
& \textbf{28.04} & \textbf{28.58} & \textbf{29.74}& \textbf{0.1917} & \textbf{0.5946} & \textbf{14.312} & \textbf{0.8010} & \textbf{0.3019}
& \textbf{9.09} & \textbf{9.40} & \textbf{9.26} \\
\bottomrule
\end{tabular}
\label{tab:image-3d}
\end{table*}

\noindent\textbf{Baselines.} We evaluated our method against several established baselines, demonstrating strong performance across diverse frameworks. These include text-to-3D models like GaussianDreamer~\citep{yiGaussianDreamerFastGeneration2023}, Score Jacobian Chain (SJC)~\citep{wangScoreJacobianChaining2022}, DreamFusion-IF~\citep{pooleDreamFusionTextto3DUsing2022}, Magic3D~\citep{linMagic3DHighResolutionTextto3D2023}, and ProlificDreamer~\citep{wangProlificDreamerHighFidelityDiverse2023}, as well as image-to-3D models such as DreamGaussian~\citep{tangDreamGaussianGenerativeGaussian2023}, Zero-1-to-3~\citep{liuZero1to3ZeroshotOne2023}, and Wonder3D~\citep{longWonder3DSingleImage2023}. We also included methods based on NeRF~\citep{mildenhallNeRFRepresentingScenes2022} and 3DGS~\citep{kerbl3DGaussianSplatting2023} for a comprehensive comparison. 
For 4D generation, we adopt the most representative and efficient models in the field—DreamGaussian4D~\citep{renDreamGaussian4DGenerative4D2023}, 4DFY~\citep{bahmani4DfyTextto4DGeneration2024}, and Consistent4D~\citep{jiangConsistent4DConsistent360deg2023}—as our baselines, including both text-to-4D and image-to-4D.
Identical parameter configurations and seed values were maintained for fair comparison, using default hyperparameters from the baselines’ open-source implementations. We employed the Threestudio library~\citep{guoThreestudio2023} for SJC and Magic3D, and the official codebases for the other methods. 

\begin{table}[t]
\centering
\caption{Quantitative comparisons in 4D generation.}
\renewcommand{\arraystretch}{1.4}
\begin{tabular}{l ccc ccc}
\toprule
\multirow{2}{*}{\textbf{Methods}} 
& \multicolumn{3}{c}{\textbf{CLIP Scores} $\uparrow$}
& \multicolumn{3}{c}{\textbf{User Study $\uparrow$}} \\
\cmidrule(lr){2-4} \cmidrule(lr){5-7}
& \textbf{B/32} & \textbf{B/16} & \textbf{L/14} 
& \textbf{Cons.} & \textbf{Qual.} & \textbf{Align.} \\
\midrule
Cons.4D
& 22.34 & 23.05 & 23.72
& 5.17 & 6.91 & 8.60 \\
\rowcolor{blue!6}\textbf{+Hallo4D}
& \textbf{25.31} & \textbf{26.12} & \textbf{26.84}
& \textbf{8.02} & \textbf{8.65} & \textbf{9.32} \\
DG4D
& 28.62 & 29.03 & 29.87
 & 6.16 & 6.25 & 8.02 \\
\rowcolor{blue!6}\textbf{+Hallo4D}
& \textbf{32.14} & \textbf{33.02} & \textbf{33.71} 
& \textbf{8.82} & \textbf{8.94} & \textbf{9.49} \\
4DFY
& 19.83 & 20.46 & 21.07 
& 3.42 & 4.93 & 7.11 \\
\rowcolor{blue!6}\textbf{+Hallo4D}
& \textbf{28.52} & \textbf{29.21} & \textbf{29.65} 
& \textbf{8.17} & \textbf{8.51} & \textbf{9.08} \\
\bottomrule
\end{tabular}
\label{tab:4d}
\end{table}

\begin{figure*}[!t]
    \centering
    \includegraphics[width=0.95\linewidth]{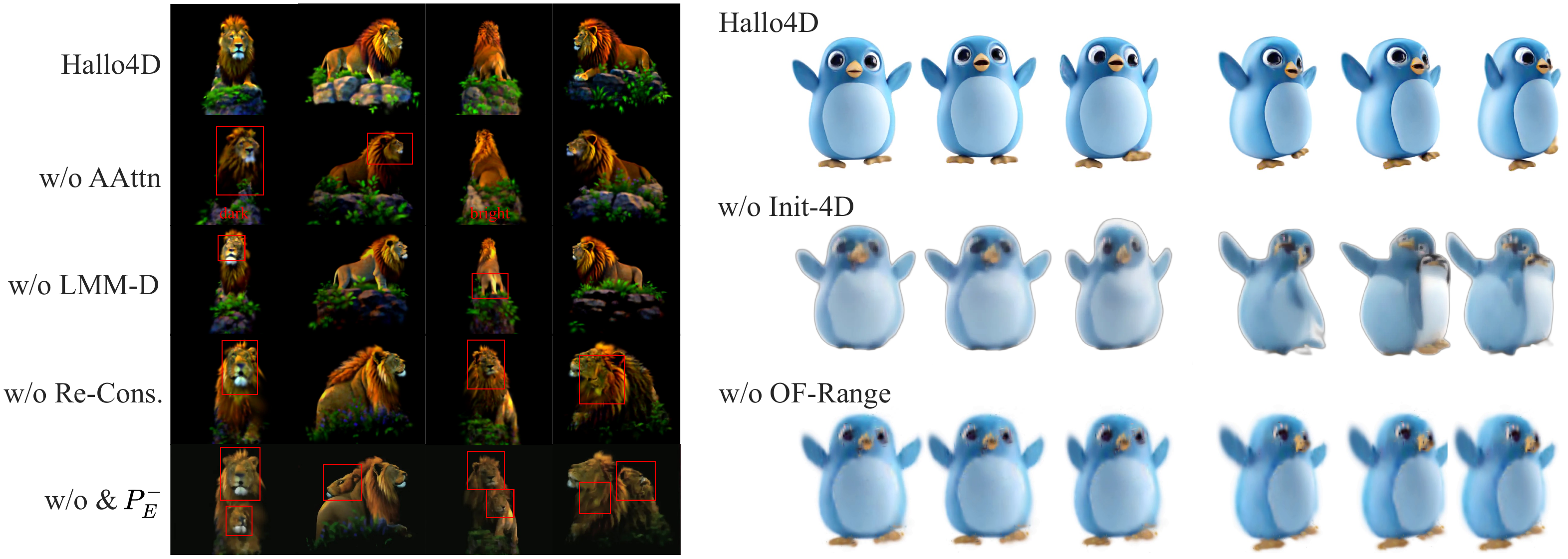}
    \caption{\textbf{Ablation study of Hallo4D.} We conduct a quantitative ablation study by successively removing each module from the full model, confirming the necessity and effectiveness of the modular design in achieving high-quality spatiotemporal generation.}
    \label{fig:ablation}
\end{figure*}

\noindent\textbf{Metrics. }In both 3D and 4D generation, the absence of ground-truth references has long hindered the development of widely accepted quantitative metrics for evaluating view consistency and spatiotemporal coherence. After reviewing a broad range of existing literature, we adopt CLIP-Score~\citep{radfordLearningTransferableVisual2021} as our primary evaluation metric. Additionally, for image-to-3D tasks, we follow the experimental setup established in prior works~\citep{longWonder3DSingleImage2023, wuUnique3DHighqualityEfficient2024}, enabled by the availability of GSO~\citep{downsGoogleScannedObjects2022} and Objaverse~\citep{deitkeObjaverseUniverseAnnotated2022} datasets. This includes evaluating geometric accuracy using Chamfer Distance (CD) and Volume IoU (Vol. IoU) between ground-truth and reconstructed shapes, and assessing visual quality using PSNR, SSIM~\citep{zhouwangImageQualityAssessment2004}, and LPIPS~\citep{zhangUnreasonableEffectivenessDeep2018a}.

\subsection{Qualitative Comparison with Baselines}
We perform a comprehensive qualitative comparison between Hallo4D and a range of baseline methods across three major tasks: text-to-3D (Fig.~\ref{fig:expr_text_to_3D}), image-to-3D (Fig.~\ref{fig:expr_image_to_3D}), and 4D generation (Fig.~\ref{fig:expr_4D}). For each task, we directly compare the original outputs of baseline models with the corresponding outputs after applying our Hallo4D optimization. To ensure a fair evaluation, we do not modify the original configurations or generation procedures of the baselines. 
Across all scenarios, Hallo4D demonstrates substantial improvements in both geometric fidelity and consistency. In 3D tasks, it produces sharper textures, more stable structures, and notably mitigates typical Janus artifacts such as multi-head or extra-limb distortions. In 4D generation, Hallo4D achieves higher spatial-temporal coherence with fewer frame-to-frame inconsistencies or identity flickers. Notably, for text-driven 4D generation, Hallo4D shows clear advantages over 4D-FY, significantly reducing geometry collapse and temporal drift. These results highlight the robustness and generalizability of Hallo4D across diverse generative conditions and modalities.

\subsection{Quantitative Comparison with Baselines}
\textbf{Computational results. }
Following~\citep{yiGaussianDreamerFastGeneration2023, qiuRichDreamerGeneralizableNormalDepth2023, tangDreamGaussianGenerativeGaussian2023}, we generated 80 unique 3D prompts using ChatGPT~\citep{openaiChatGPT} and arranged 16 cameras uniformly in a 360-degree configuration around the z-axis to evaluate the CLIP-Score. For 3D generation, the average CLIP-Score across all views was used to evaluate multi-view consistency, while for 4D generation, we computed the average CLIP-Score across all frames to assess spatiotemporal consistency. 
For image-conditioned models, we selected 60 objects from the GSO~\citep{downsGoogleScannedObjects2022} and Objaverse~\citep{deitkeObjaverseUniverseAnnotated2022}, replacing overly simplistic instances to ensure a more challenging and comprehensive evaluation. Each object was rendered from a frontal view at a resolution of 256×256 and used as input for computing CD, Vol. IoU, PSNR, SSIM~\citep{zhouwangImageQualityAssessment2004}, and LPIPS~\citep{zhangUnreasonableEffectivenessDeep2018a} to assess both geometric accuracy and visual quality. It should be noted that the existence of a ground truth corresponding to the front view in image-driven generation generally leads to higher generation quality. 
The quantitative results for text-to-3D, image-to-3D, and 4D generation are presented in Tab.~\ref{tab:text-3d}, Tab.~\ref{tab:image-3d}, and Tab.~\ref{tab:4d}, respectively. The results show that Hallo4D consistently outperforms all baselines across all evaluation metrics, demonstrating its effectiveness in enhancing spatiotemporal consistency for both 3D and 4D generation. 

\begin{table}[t]
\centering
\caption{Quantitative ablation study.}
\renewcommand{\arraystretch}{1.4}
\setlength{\tabcolsep}{5.6pt}
\resizebox{\columnwidth}{!}{%
\begin{tabular}{l ccc ccc}
\toprule
\multirow{2}{*}{\textbf{Modules}}
& \multicolumn{3}{c}{\textbf{CLIP Scores} $\uparrow$}
& \multicolumn{3}{c}{\textbf{User Study $\uparrow$}} \\
\cmidrule(lr){2-4} \cmidrule(lr){5-7}
& \textbf{B/32} & \textbf{B/16} & \textbf{L/14} 
& \textbf{Cons.} & \textbf{Qual.} & \textbf{Align.} \\
\midrule
3D Baseline
& 21.27 & 22.67 & 23.71
& 6.00 & 5.53 & 5.57 \\
w/o AAttn
& 23.98 & 25.88 & 29.36
& 7.64 & 7.79 & 7.93 \\
w/o LMM-D
& 23.65 & 25.10 & 28.71 
& 7.25 & 7.32 & 7.53 \\
w/o Re-Cons.
& 22.46 & 23.59 & 26.92 
& 6.83 & 7.01 & 6.80 \\
w/o \& $P^-_E$ 
& 22.23 & 23.23 & 25.58 
& 6.54 & 6.69 & 6.43 \\
\rowcolor{blue!7}\textbf{Hallo4D}
& \textbf{24.45} & \textbf{26.83} & \textbf{30.00} 
& \textbf{8.87} & \textbf{8.67} & \textbf{8.87} \\
\midrule
4D Baseline
& 28.62 & 29.03 & 29.87
& 6.16 & 6.25 & 8.02 \\
w/o Init-4D
& 30.41 & 30.92 & 31.40
& 7.04 & 7.33 & 8.52 \\
w/o OF-Range
& 29.28 & 29.82 & 30.36 
& 6.91 & 7.21 & 8.40 \\
\rowcolor{blue!4} w/o CESA \& LDR
& 31.88 & 32.65 & 33.34 
& 8.50 & 8.65 & 9.21 \\
\rowcolor{blue!9} w/o Consensus
& 31.95 & 32.71 & 33.39 
& 8.58 & 8.70 & 9.29 \\
\rowcolor{blue!17}\textbf{Hallo4D}
& \textbf{32.14} & \textbf{33.02} & \textbf{33.71} 
& \textbf{8.82} & \textbf{8.94} & \textbf{9.49} \\
\bottomrule
\end{tabular}%
}
\label{tab:ablation}
\end{table}

\textbf{User study. }We conducted a user study with 58 volunteers specializing in artificial intelligence to evaluate the quality of generated 3D and 4D content. To facilitate comprehensive comparison, participants were asked to rate each model on three criteria—“Multi-view Consistency,” “Overall Quality,” and “Alignment with Prompt”—using a 10-point scale. For 3D generation, each asset was rendered at 15° intervals and compiled into a video to visualize view consistency. For 4D generation, we incorporated continuous camera rotation during playback, allowing simultaneous observation of spatial and temporal changes. This design offers an intuitive and immersive way to assess spatiotemporal consistency. Final scores were obtained by averaging user ratings, as shown in the last columns of Tab.~\ref{tab:text-3d}, Tab.~\ref{tab:image-3d}, and Tab.~\ref{tab:4d}. The results demonstrate that Hallo4D achieves significant improvements over baseline models, further validating the effectiveness of our approach.

\subsection{Ablation Study}
We conducted ablation experiments on the individual modules of Hallo4D, as shown in Fig.~\ref{fig:ablation}. Starting from the full model, we removed each module independently to evaluate its impact. For clarity, we denote \textit{AAttn} as Multi-view Appearance Alignment, \textit{LMM-D} as Multi-modal Hallucination Detection, \textit{Re-Cons.} as Prompt-Enhanced Re-Consistency and \textit{Init-4D} as Initialization for 4D Optimization. Notably, for 3D generation under the “w/o Re-Cons.” setting, the LMM output $P_E^{-}$ is still applied in $\mathcal{L}_{\rm{SDS}}$ to demonstrate the necessity of $\mathcal{L}_{\rm{CG}}$. We further include a setting “w/o \& $P_E^{-}$”, where $P_E^{-}$ is not used at all, to more clearly isolate and verify the effectiveness of the \textit{Re-Cons.} module.

In Fig.~\ref{fig:ablation}, we evaluate the contribution of each module. Removing \textit{AAttn} leads to noticeable brightness imbalance and unnatural color shifts, which affect texture fidelity. Modules \textit{LMM-D} and \textit{Re-Cons.} are critical for ensuring cross-view consistency: without \textit{LMM-D}, the lion’s head is visibly deformed or missing across views; without \textit{Re-Cons.}, duplicated facial features and extra limbs appear. When both \textit{Re-Cons.} and $P_E^{-}$ are removed (“w/o \& $P_E^{-}$”), these artifacts become more pronounced, reinforcing the necessity of LMM-guided correction.
The “w/o Init-4D” setting results in significant degradation of both spatial and temporal consistency in 4D generation, highlighting the crucial role of initialization in improving early-stage optimization. Additionally, removing module \textit{OF-Range} weakens the impact of motion cues from the animated video, leading to reduced temporal alignment.

Furthermore, we conduct a comprehensive ablation study to validate the effectiveness of the proposed modules. Specifically, \textit{CESA \& LDR} denotes the Adaptive Exposure-Aware Semantic Alignment, and \textit{Consensus} stands for the LMM Consensus Selector. Note that we adopt an incremental ablation setup; thus, the baseline configuration excluding \textit{CESA \& LDR} inherently omits the \textit{Consensus} module as well. The ablation results demonstrate that the sequential integration of each component steadily improves the model's performance in 4D generation. These modules are critical for enhancing the overall generation quality, as they offer targeted strategies to resolve distinct technical challenges.
To complement these qualitative findings, Tab.~\ref{tab:ablation} presents quantitative results that further confirm the necessity and effectiveness of each module in the Hallo4D framework.

\section{Conclusion}
In this work, we propose Hallo4D, a unified and model-agnostic framework for mitigating spatiotemporal hallucinations in both 3D and 4D content generation. By leveraging the spatial and temporal reasoning capabilities of LMMs, our generation-detection-correction paradigm identifies inconsistencies across views and frames and guides image-space optimization accordingly. Crucially, to ensure robust geometric fidelity and prevent compounding errors inherent in single-pass edits, these insights guide a consensus-driven image-space optimization mechanism. This approach utilizes an MLLM Consensus Selector to evaluate multiple candidate corrections via multi-model voting. To further enhance 4D consistency, Hallo4D incorporates an optical flow-based keyframe sampling strategy, an LMM-guided initialization scheme, and an attention-based appearance alignment module. To address exposure instability, we introduce two complementary losses—Contrastive Semantic Exposure Alignment (CSEA) and a log-dynamic-range (LDR) term—and apply union-of-frusta visibility pruning to suppress out-of-view clutter and reduce pseudo under-exposure. Without requiring retraining or architectural changes, our method improves generation quality across spatial and temporal dimensions. Extensive experiments confirm that Hallo4D consistently outperforms strong baselines, offering a scalable and effective solution for consistency-aware 3D and 4D generation.



%
%

\bibliographystyle{spbasic}
\bibliography{reference}

\end{sloppypar}

\end{document}